\documentclass[]{style}

\definecolor{lightblue}{rgb}{0.0, 0.6, 1.0}
\definecolor{darkgreen}{rgb}{0.0, 0.6, 0.0}
\definecolor{lightgreen}{rgb}{0.1, 0.8, 0.1}
\definecolor{lightred}{rgb}{0.7, 0.7, 0.7}
\definecolor{lightgreen}{rgb}{0, 0, 0}
\definecolor{lightlightgray}{rgb}{0.8, 0.8, 0.8}

\newcommand{\red}[1]{\textcolor{red}{#1}}
\newcommand{\blue}[1]{\textcolor{blue}{#1}}

\newcommand{\colordown}[1][]{\red{$\downarrow$#1}}
\newcommand{\colorup}[1][]{\blue{$\uparrow$#1}}

\usepackage{titlesec}
\titlespacing*{\paragraph}
{0pt}{0em}{0.2em}  

\makeatletter
\renewcommand\paragraph{\@startsection{paragraph}{4}{\z@}%
  {3pt}
  {-0.5em}
  {\normalfont\normalsize\bfseries}} 
\makeatother

\usepackage{amsmath}  
\usepackage{upgreek}
\usepackage{array}    
\usepackage{multirow}
\usepackage{array}
\usepackage{algorithm}
\usepackage{makecell}
\usepackage{graphicx}
\usepackage{algorithm}
\usepackage{algpseudocode}
\usepackage{colortbl}
\usepackage{enumitem}
\usepackage{wrapfig}
\usepackage{mathrsfs}
\usepackage{pifont}
\usepackage{soul}
\usepackage{amssymb}
\usepackage{tcolorbox}
\usepackage{dirtytalk}
\usepackage{xspace}
\usepackage[utf8]{inputenc}

\usepackage{booktabs}

\usepackage{url}
\definecolor{blue}{rgb}{0.21,0.49,0.74}
\definecolor{red}{rgb}{0.8, 0.2, 0.2}
\definecolor{green}{rgb}{0, 0.5, 0}
\definecolor{yellow}{RGB}{218, 160, 109}
\definecolor{gray}{RGB}{155, 155, 155}

\crefname{section}{Sec.}{Secs.}
\Crefname{section}{Section}{Sections}
\Crefname{table}{Table}{Tables}
\crefname{table}{Tab.}{Tabs.}
\crefname{figure}{Fig.}{Figs.}
\Crefname{figure}{Figure}{Figures}
\crefname{appendix}{App.}{Apps.}
\Crefname{appendix}{Appendix}{Appendices}

\newcommand{\OMs}{\texttt{PISCO-1.3B}\xspace}
\newcommand{\OMl}{\texttt{PISCO-14B}\xspace}
\newcommand{\OM}{\texttt{PISCO}\xspace}
\newcommand{\OMb}{\texttt{PISCO-Bench}\xspace}

\usepackage[utf8]{inputenc} 
\usepackage[T1]{fontenc}    
\usepackage{graphicx}
\usepackage{amsmath}
\usepackage{amssymb}
\usepackage{url}            
\usepackage{booktabs}       
\usepackage{amsfonts}       
\usepackage{nicefrac}       
\usepackage{microtype}      
\usepackage{xcolor}         
\usepackage{multirow}
\usepackage{comment}
\usepackage{colortbl}
\usepackage{tocloft}
\usepackage{algorithm}
\usepackage{algorithmicx}
\usepackage{algpseudocode}
\usepackage{wrapfig}
\usepackage{tabularx}
\usepackage{textcomp}
\usepackage{stfloats}
\usepackage{verbatim}
\usepackage{titlesec}
\usepackage{adjustbox}
\usepackage{pifont}
\usepackage{tikz}
\usepackage{color}
\usepackage{subcaption}
\usepackage{soul}

\RequirePackage{xspace}
\makeatletter
\DeclareRobustCommand\onedot{\futurelet\@let@token\@onedot}
\def\@onedot{\ifx\@let@token.\else.\null\fi\xspace}

\makeatother

\definecolor{lightblue}{rgb}{0.66, 0.85, 0.95}
\hypersetup{
    colorlinks=true,
    linkcolor=red,
    citecolor=blue,
}
\definecolor{c2}{HTML}{FBD9BD}
\definecolor{c3}{HTML}{fe793d}
\definecolor{c4}{HTML}{eedeb0}
\definecolor{rouse}{rgb}{0.981,0.961,0.941}
\usepackage[most]{tcolorbox}
\tcbset{colback=blue!5!white, colframe=blue!20!white, boxrule=0pt, arc=2mm, left=1mm, right=1mm, top=1mm, bottom=1mm}

\definecolor{adptorange}{RGB}{248, 205, 172}
\definecolor{cmpblue}{RGB}{189, 215, 238}
\definecolor{cmpblue}{RGB}{189, 215, 238}

\definecolor{our_red}{RGB}{232,157,160}
\definecolor{our_blue}{RGB}{136,206,230}
\definecolor{our_orange}{RGB}{246,200,168}
\definecolor{our_green}{RGB}{178,211,164}

\definecolor{attn_code0}{RGB}{247,215,200}
\definecolor{attn_code1}{RGB}{238,169,139}
\definecolor{mlp_code0}{RGB}{204,201,221}
\definecolor{mlp_code1}{RGB}{102,95,153}

\definecolor{token_blue}{RGB}{84, 120, 140}
\def\blue#1{\textbf{\color{our_blue} #1}} 
\def\red#1{\textbf{\color{our_red} #1}} 

\usepackage{pifont}       
\usepackage{bbding}       
\usepackage{fontawesome}
\usepackage{xspace}

\newcommand{\cmark}{\ding{51}}
\usepackage{float}

\newlength\savewidth

\newcolumntype{x}[1]{>{\centering\arraybackslash}p{#1pt}}
\newcolumntype{y}[1]{>{\raggedright\arraybackslash}p{#1pt}}
\newcolumntype{z}[1]{>{\raggedleft\arraybackslash}p{#1pt}}

\renewcommand{\paragraph}[1]{\vspace{1mm}\noindent\textbf{#1}}

\usepackage{colortbl}
\usepackage{xcolor}
\usepackage{wrapfig}

\renewcommand{\paragraph}[1]{\vspace{1.25mm}\noindent\textbf{#1}}

\usepackage{listings}

\definecolor{codeblue}{rgb}{0.21, 0.49, 0.74}
\definecolor{codekw}{rgb}{0.35, 0.35, 0.75}
\lstdefinestyle{Pytorch}{
    language = Python,
    backgroundcolor = \color{white},
    basicstyle = \fontsize{9pt}{8pt}\selectfont\ttfamily\bfseries,
    columns = fullflexible,
    aboveskip=1pt,
    belowskip=1pt,
    breaklines = true,
    captionpos = b,
    commentstyle = \color{codeblue},
    keywordstyle = \color{codekw},
}

\definecolor{green}{HTML}{009000}
\definecolor{red}{HTML}{ea4335}

\title{PISCO: \underline{P}recise Video Instance \underline{I}nsertion \\ with \underline{S}parse \underline{Co}ntrol}
\author[1]{Xiangbo Gao}
\author[1]{Renjie Li}
\author[1]{Xinghao Chen}
\author[3]{Yuheng Wu}
\author[4]{Suofei Feng}
\author[2]{Qing Yin}
\author[1,2]{Zhengzhong Tu}

\affiliation[1]{Texas A\&M University}
\affiliation[2]{Visko Platform}
\affiliation[3]{KAIST}
\affiliation[4]{Stanford University}

\abstract{
The landscape of AI video generation is undergoing a pivotal shift: moving beyond general generation - which relies on exhaustive prompt-engineering and "cherry-picking" - towards fine-grained, controllable generation and high-fidelity post-processing. In professional AI-assisted filmmaking, the core requirement is the ability to perform precise, targeted modifications. A cornerstone of this transition is video instance insertion, which requires inserting a specific instance into existing footage while maintaining scene integrity. Unlike traditional video editing, this task demands several requirements: precise spatial-temporal placement, physically consistent scene interaction (e.g., shadows and reflections), and the faithful preservation of original dynamics - all achieved under minimal user effort. In this paper, we propose \OM, a video diffusion model for precise video instance insertion with arbitrary sparse keyframe control. \OM allows users to specify a single keyframe, start-and-end keyframes, or sparse keyframes at arbitrary timestamps, and automatically propagates object appearance, motion, and interaction. To address the severe distribution shift induced by sparse conditioning in pretrained video diffusion models, we introduce Variable-Information Guidance for robust conditioning and Distribution-Preserving Temporal Masking to stabilize temporal generation, together with geometry-aware conditioning for realistic scene adaptation. We further construct \OMb, a benchmark with verified instance annotations and paired clean background videos, and evaluate performance using both reference-based and reference-free perceptual metrics. Experiments demonstrate that \OM consistently outperforms strong inpainting and video editing baselines under sparse control, and exhibits clear, monotonic performance improvements as additional control signals are provided.
}

\project{\url{https://xiangbogaobarry.github.io/PISCO/}}
\date{\today}
\contact{Xiangbo Gao (\href{mailto:xiangbogaobarry@gmail.com}{xiangbogaobarry@gmail.com}), Zhengzhong Tu (\href{mailto:tzz@tamu.edu}{tzz@tamu.edu})}

\begin{document}
\thispagestyle{firstheader}
\maketitle
\pagestyle{plain}

\begin{figure*}[t]
\centering
\includegraphics[width=\linewidth]{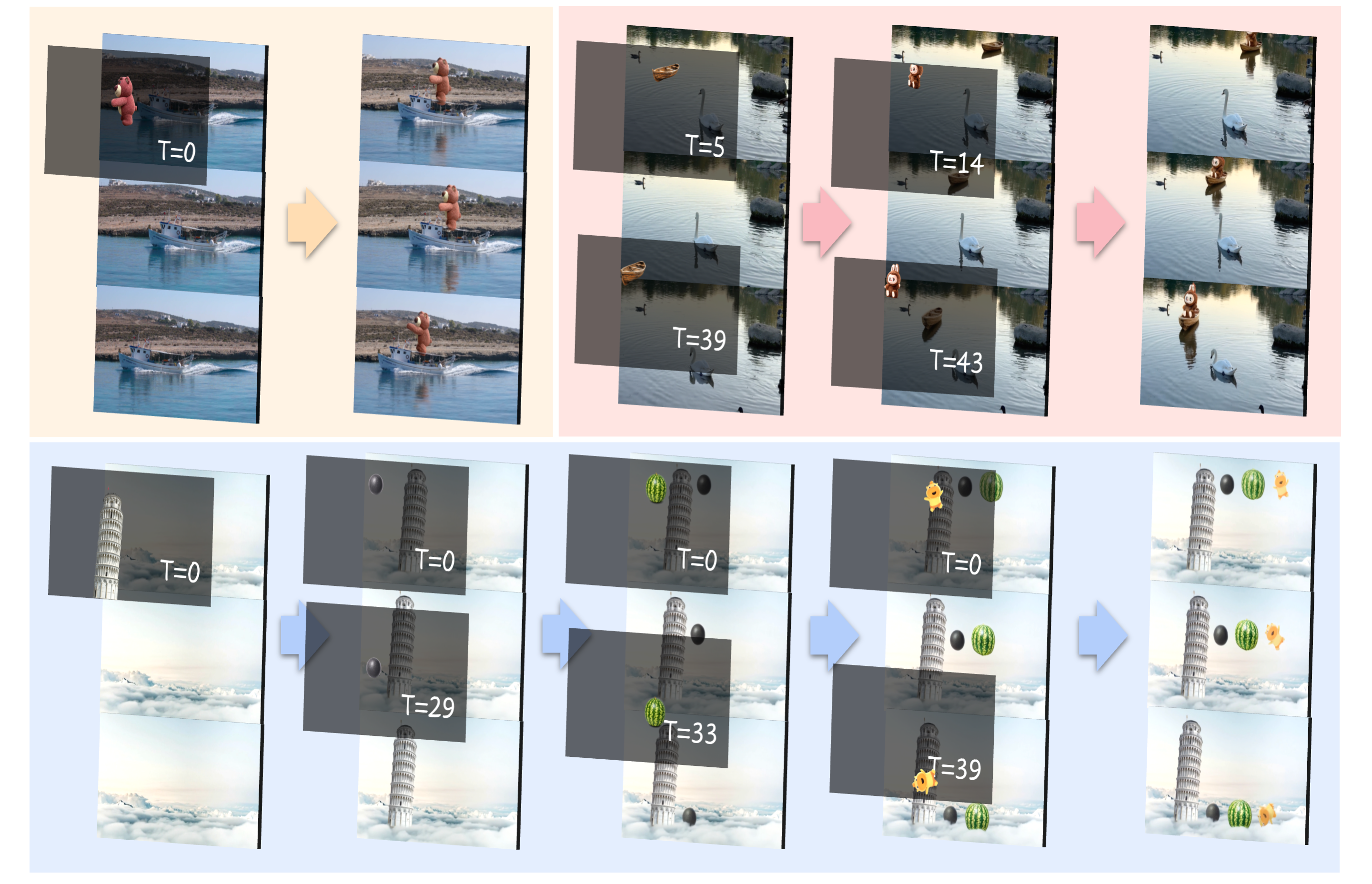}
\caption{\OM enables precise video instance insertion with arbitrary sparse keyframe control. Given a clean input video and a few user-provided instance cutouts at selected timestamps, \OM inserts the instance with coherent temporal propagation and physical effects while preserving the original background dynamics.}
\label{fig:datademo}
\vspace{-0.3cm}
\end{figure*}

\section{Introduction}

Recent advances in large-scale video generative models~\cite{kong2024hunyuanvideo, wan2025wan, hacohen2026ltx} are able to generate videos with high visual fidelity and realistic motion. 
As these models continue to evolve, the focus of video generation is shifting beyond producing visually plausible content toward highly controllable video generation and editing~\cite{polyak2024movie,gao2026viskoorbis,wu2026deltaforcing}, with the long-term goal of enabling \textbf{Hollywood-grade AI-assisted filmmaking}~\cite{manovich2023generative,moore2024video}. In this new paradigm, the priority is no longer to \textbf{cherry-pick} a barely acceptable output from numerous generations, but to reliably \textbf{achieve precise user intent} with minimal iteration.

A particularly demanding yet under-explored capability in this context is
\textbf{Precise Video Instance Insertion}: the ability to insert a specific object into an existing video at a user-specified spatial location and temporal position, while preserving the identity and dynamics of the original footage.
In professional visual effects and post-production workflows~\cite{gao2026vefxbench}, instance insertion is not merely about adding a visually plausible object, but about achieving precise and reliable control with minimal iteration.
In this setting, precision entails several tightly coupled requirements:
\begin{itemize}[leftmargin=5mm, nosep]
\item[\ding{202}] \textbf{Instance-level controllability}, enabling users to explicitly specify when and where an object appears;
\item[\ding{203}] \textbf{Physically plausible temporal propagation}, where the inserted object automatically evolves over time with coherent pose and motion, following physically reasonable dynamics;
\item[\ding{204}] \textbf{Physically consistent scene adaptation}, in which the surrounding background is appropriately adjusted to account for insertion-induced effects such as shadows, reflections, illumination, water ripples, and others;
\item[\ding{205}] \textbf{Faithful preservation of background scene dynamics}, ensuring that pre-existing motions, identities, and temporal patterns in the original video remain unchanged and temporally consistent after insertion;
\item[\ding{206}] \textbf{Low-effort user interaction}~\cite{yu2026sparkvsr}, where achieving the above goals does not require dense per-frame annotations or manual editing across the entire video.
\end{itemize}

Despite the strong generative capacity of diffusion-based video models~\cite{yang2023diffusion}, these requirements remain largely unmet.
Existing solutions can be broadly categorized into several classes, each addressing only a subset of the above requirements.
Video inpainting methods~\cite{quan2024deep, zi2025cococo} enforce spatial consistency through dense per-frame masks, but require exhaustive annotation of object shape and trajectory, making them impractical under sparse user control (\ding{206}).
Moreover, their copy-and-fill nature limits the ability to adjust the surrounding scene in a physically consistent manner, such as modeling realistic shadows or illumination changes (\ding{204}).
Reference-guided video-to-video editing approaches~\cite{wei2025univideo, jiang2025vace} propagate appearance information over time, yet lack fine-grained spatial and temporal controllability, making it difficult to precisely specify object placement and timing (\ding{202}).
Agentic pipelines that combine image editing with subsequent image-to-video generation~\cite{hu2022make, jin2024pyramidal} allow flexible object manipulation at the image level, but discard most temporal information from the original video, often leading to background drift and disrupted scene dynamics (\ding{205}).
Geometry-heavy reconstruction-based approaches~\cite{liu2024place, jin2025insertanywhere} explicitly model scene structure and spatial relationships, but incur substantial computational overhead, rely on fragile geometric cues, and struggle to robustly model complex or previously unseen object motions (\ding{203}).
Together, these limitations prevent existing methods from providing a practical and flexible solution for precise video instance insertion under sparse user control.

To overcome these limitations, we propose \OM (\underline{P}recise \underline{I}nstance insertion with \underline{S}parse \underline{CO}ntrol), a video diffusion framework designed for professional-grade instance insertion with minimal user effort. \OM allows users to specify sparse instance conditions—such as a single keyframe, start-and-end keyframes, or arbitrary keyframes at any timestamps—and performs video instance insertion by propagating appearance, motion, and interaction in a scene-consistent manner. Our framework builds upon the Wan video diffusion backbone and augments it with a multi-channel context adapter that ingests instance RGB, mask, depth, and an explicit availability signal indicating when instance guidance is provided, enabling flexible sparse keyframe control within a unified model.

Sparse control introduces additional challenges, as na\"ively masking missing frames can cause severe distribution shift when processed by pretrained temporal video VAEs, resulting in flickering, miscoloring, and incomplete instance rendering. \OM addresses these challenges through a set of dedicated mechanisms, including \textbf{Variable-Information Guidance (VIG)} for modulating conditioning strength during training to enable robust guidance, \textbf{Distribution-Preserving Temporal Masking (DPTM)} to stabilize sparse conditioning under temporal VAEs, and geometry-aware conditioning to maintain physically plausible interactions and scene consistency.

To evaluate precise video instance insertion under sparse control, we construct \OMb, a curated benchmark derived from BURST~\cite{athar2023burst} with carefully verified instance annotations and paired clean background videos generated using a side-effect-aware instance removal model~\cite{miao2025rose}. We evaluate performance using both reference-based metrics (FVD, LPIPS, PSNR, SSIM) and reference-free perceptual metrics (VBench~\cite{huang2024vbench}). Experimental results show that \OM consistently outperforms strong inpainting and video editing baselines under sparse-control settings, and further demonstrates clear and monotonic improvements as additional sparse control frames are provided, validating its scalability with respect to control signal density. Our contributions are summarized as follows:
\begin{itemize}[leftmargin=*, nosep]
    \item We introduce \OM, the first video diffusion framework that enables precise video instance insertion under arbitrary sparse keyframe control, allowing users to insert instances at any desired timestamps with minimal annotation effort.
    \item We propose a set of dedicated mechanisms for sparse instance control, including Variable-Information Guidance and Distribution-Preserving Temporal Masking, which jointly address temporal propagation and distribution shift in pretrained video diffusion models.
    \item We construct \OMb and demonstrate through extensive reference-based and reference-free experiments that \OM significantly outperforms state-of-the-art video instance insertion baselines, exhibiting consistent performance improvements as additional control signals are provided.
\end{itemize}


\section{Related Works}

\subsection{Video Inpainting for Video Instance Insertion}
Video inpainting~\cite{quan2024deep} refers to masking out a certain part of the video and leveraging models to recover the masked region. It serves as a core technique for object removal~\cite{li2026physicsaware}, insertion, replacement, video restoration, and outpainting. Early methods employed 3D Convolutional Neural Networks (3D-CNNs) to learn spatio-temporal coherence~\cite{wang2019video, chang2019free, hu2020proposal}. To provide stronger motion priors, subsequent approaches incorporated optical flow guidance~\cite{xu2019deep, kim2019deep, li2020short, gao2020flow, zou2021progressive, li2022towards}. With the advent of Transformers, attention mechanisms were adopted to better capture long-range spatio-temporal correlations~\cite{liu2021fuseformer, liu2021decoupled, cai2022devit, zhou2023propainter}.

Recently, with the development of diffusion models~\cite{yang2023diffusion} and video generative foundation models (e.g., Wan~\cite{wan2025wan}, Flux~\cite{labs2025flux}), video diffusion models have been extensively applied to video inpainting~\cite{guo2023animatediff, zhang2024avid, green2024semantically, zi2025cococo, bian2025videopainter, li2025diffueraser, lee2025video, wan2025unipaint, guo2025keyframe}. A significant advantage of these models is that training data can be obtained by simply masking out target regions in original videos, relieving the reliance on high-quality paired data which is difficult to acquire. However, despite their generation quality, applying standard inpainting to instance insertion faces a critical bottleneck: it necessitates \textbf{dense segmentation annotations (masks) for every frame} during inference. Obtaining per-frame masks for a dynamic object that does not yet exist in the video is labor-intensive and impractical for general users. In contrast, our proposed method requires only \textbf{sparse instance mask signals} to control the video instance insertion process.

\subsection{Reference-Guided Video-to-Video Editing}
Reference-guided video editing focuses on conditional video generation, using references such as instance images or text prompts to guide the synthesis process. Large-scale frameworks like UniVideo~\cite{wei2025univideo} and VACE~\cite{jiang2025vace} integrate multimodal controls to unify generation and editing tasks. Similarly, methods such as VideoDirector~\cite{wang2025videodirector} and ContextFlow~\cite{chen2025contextflow} utilize text-to-video models to modify scene content via prompt engineering. Specific to object manipulation, VideoAnydoor~\cite{tu2025videoanydoorhighfidelityvideoobject} and Wan-Animate~\cite{cheng2025wan} focus on high-fidelity object insertion or replacement with motion control, while Animate-a-Story~\cite{he2023animate} retrieves assets for storytelling.

The primary limitation of these methods is the inability to achieve \textbf{fine-grained spatial and temporal control}. Most current approaches function as general video-to-video translation, which modifies the global style or replaces existing objects where the motion trajectory is pre-defined. They struggle to satisfy precise user intents, such as inserting a specific instance at an \textit{exact position}, with an \textit{exact posture}, starting at an \textit{exact timestamp}. Our work specifically addresses this need, enabling precise instance-level manipulation while maintaining the integrity of the original video.

\subsection{Image-to-Video (I2V) Generation and Propagation}
A distinct paradigm involves a ``generate-then-animate'' workflow: editing a single keyframe using mature image editing tools and then animating it via Image-to-Video (I2V) models~\cite{wu2026vgbeconsistent}. Recent advancements in image editing, such as Step1x-edit~\cite{liu2025step1x}, Z-Image~\cite{cai2025z}, and Qwen-Image~\cite{wu2025qwen}, allow for precise object insertion in static images. To extend this to video, I2V models~\cite{hu2022make, namekata2024sg, wang2024generative, jin2024pyramidal, li2025anyi2v,wu2026consid} and transition generation methods~\cite{yang2025versatile, wang2025tip} predict future frames based on the edited image. Specialized architectures like MotionStone~\cite{shi2025motionstone}, MotionPro~\cite{zhang2025motionpro}, and Through-The-Mask~\cite{yariv2025through} attempt to decouple motion from appearance to improve controllability.

However, relying solely on I2V models for insertion introduces a fundamental issue: \textbf{background hallucination}. I2V models are trained to generate motion for the entire frame and often fail to strictly adhere to the unedited background pixels of the original video. This leads to severe temporal drift and inconsistency between the edited foreground and the static background. While Generative Video Propagation~\cite{liu2025generative} attempts to mitigate this, ensuring seamless blending without altering the non-object regions remains a significant challenge for pure I2V approaches. Our approach overcomes this by leveraging the original video context to strictly preserve background consistency.

\subsection{4D Reconstruction and Manipulation}
To address challenges related to occlusion and geometric consistency, recent works have proposed lifting 2D video into explicit 3D or 4D representations~\cite{li2025fourkfourgen}. Methods like ``Place Anything into Any Video''~\cite{liu2024place} and ``Anything in Any Scene''~\cite{bai2024anything} utilize depth and flow to physically place objects. Most notably, InsertAnywhere~\cite{jin2025insertanywhere} bridges 4D scene geometry with diffusion models to handle complex occlusions during insertion. 

Despite their theoretical correctness, these geometry-heavy approaches suffer from inherent flexibility limitations compared to 2D generative models. First, they are constrained by \textbf{data scarcity}, as high-quality 4D-labeled data is rare, leading to poor generalization across diverse real-world scenes. Second, they struggle with \textbf{dynamic object insertion}; inserting a moving character into a 4D scene typically requires a pre-existing dynamic object, whereas 2D diffusion models can infer realistic motion from a single or a few images. Our method bypasses these heavy reconstruction requirements, offering a flexible, data-efficient solution that supports dynamic object insertion with natural interaction.

\section{Methodology}

\subsection{Problem Formulation}

Given paired videos $\{\hat{V}, V\}$ of length $T$, where $\hat{V}=\{\hat{V}_t\}_{t=1}^{T}$ contains a foreground instance and $V=\{V_t\}_{t=1}^{T}$ depicts the same scene without that instance, our goal is to synthesize an edited video $\tilde{V}$ in which the instance is naturally inserted into $V$ with spatially accurate placement, temporally coherent motion, and geometrically plausible occlusion relationships.

We represent the instance-side conditions as an RGB instance clip $I=\{I_t\}_{t=1}^{T}$, its spatial mask $M=\{M_t\}_{t=1}^{T}$, and the instance depth $D_I=\{D_{I,t}\}_{t=1}^{T}$. Here, each $I_t$ contains only the segmented foreground instance on a constant zero-valued background, and $M_t$ indicates the corresponding foreground region. In addition, we compute a background depth map $D_V$ from $V$ to provide geometric cues for depth ordering and occlusion reasoning. Details of how these signals are constructed are described in Sec.~\ref{sec:impl_data}.

\paragraph{Variable-density user guidance via availability mask.}
User guidance can range from a single keyframe to dense per-frame supervision. We model this variability with an availability mask $A=\{A_t\}_{t=1}^{T}$, where $A_t\in\{0,1\}$ indicates whether instance-side information is available at time $t$. During training, we sample $A \sim p_\gamma(A)$, where the density ratio $\gamma\in[0,1]$ controls the expected fraction of available frames as well as their temporal placement. The mask is applied exclusively to instance-specific signals:
\begin{equation}
    I^{A} = A \odot I, \quad
    M^{A} = A \odot M, \quad
    D_I^{A} = A \odot D_I,
\end{equation}
while the background depth $D_V$ remains unmasked. This design reflects practical scenarios: while a user typically specifies instance-level details sparsely, the background scene geometry is often fully observable.

\paragraph{Conditional video diffusion with context adapter.}
Let $z_0$ denote the latent representation of the target video obtained using a pretrained temporal video VAE, and let $z_t$ be the noisy latent at diffusion step $t$. We train a denoising network $\epsilon_\theta$ to predict the added noise conditioned on the background video and the masked instance-side signals, ending up with the training objective
\begin{equation}
\mathcal{L}
=
\mathbb{E}_{z_0,\epsilon,t,\,A\sim p_\gamma}
\left[
\left\|
\epsilon
-
\epsilon_\theta
\big(
z_t,\,
t,\,
V,\,
D_V,\,
I^{A},\, D_I^{A},\, M^{A},\, A
\big)
\right\|_2
\right].
\label{eq:training_objective}
\end{equation}

In our implementation, $\epsilon_\theta$ uses the Wan video diffusion backbone~\cite{wan2025wan} with a VACE context adapter~\cite{jiang2025vace} for injecting multi-channel video conditions. To accommodate our condition format, we modify the first linear layer of the VACE adapter to match the new input dimensionality, while reusing pretrained parameters for the remaining layers. Training details are provided in Sec.~\ref{sec:training_details}.


\begin{figure*}[t]
\centering
\includegraphics[width=\textwidth]{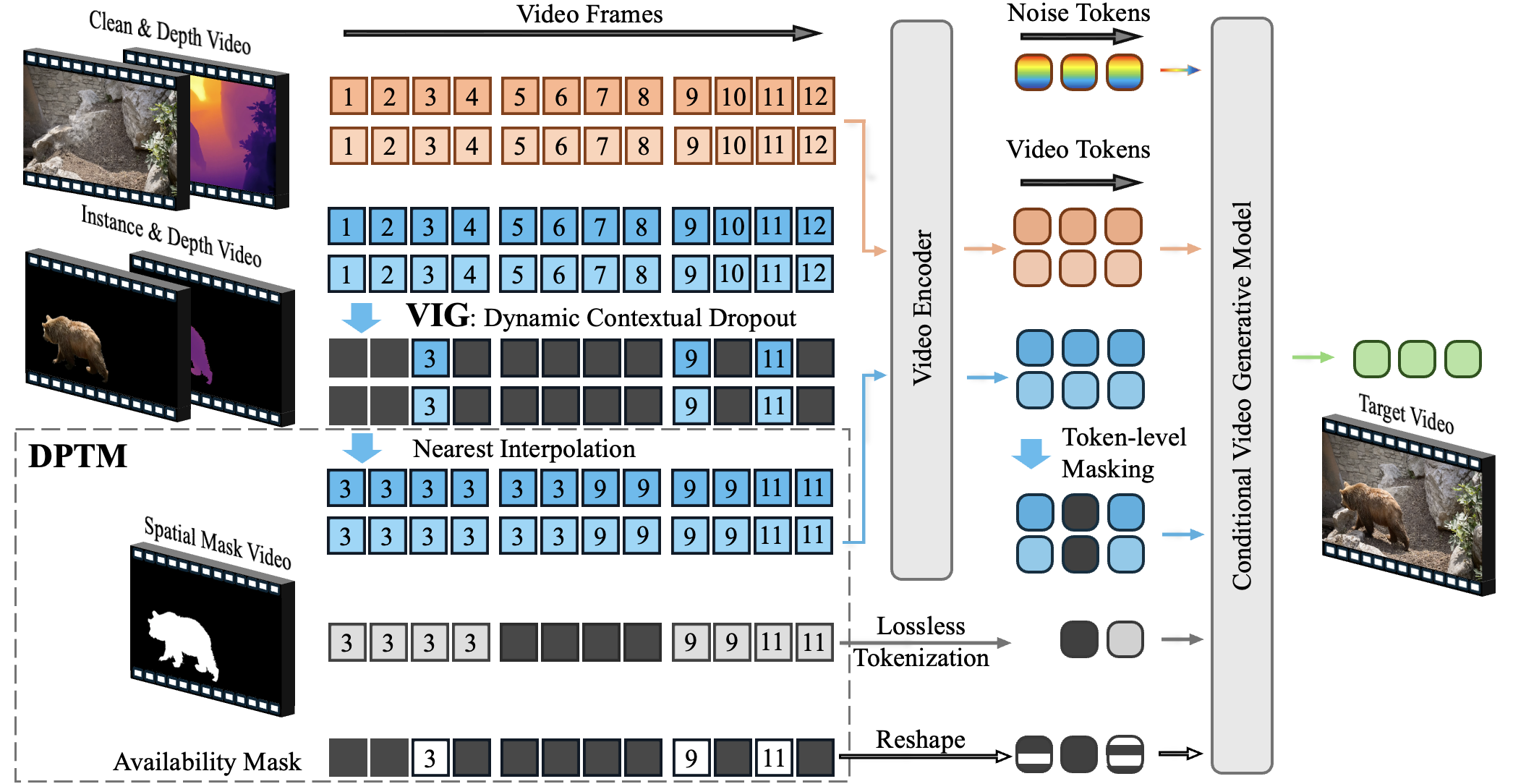}
\caption{\textbf{Overview of \OM pipeline.} We train a conditional video diffusion model with sparse keyframe control via Variable-Information Guidance (VIG), and stabilize sparse conditioning under pretrained temporal VAEs using Distribution-Preserving Temporal Masking (DPTM): pixel-space nearest-frame interpolation followed by token-level masking, with spatial mask and availability signals injected alongside RGB/depth conditions.}
\label{fig:pipeline}
\vspace{-0.2cm}
\end{figure*}

\subsection{Variable-Information Guidance via Dynamic Contextual Dropout}
\label{sec:dynamic_dropout}





\begin{figure*}[t]
\centering
\includegraphics[width=1\textwidth]{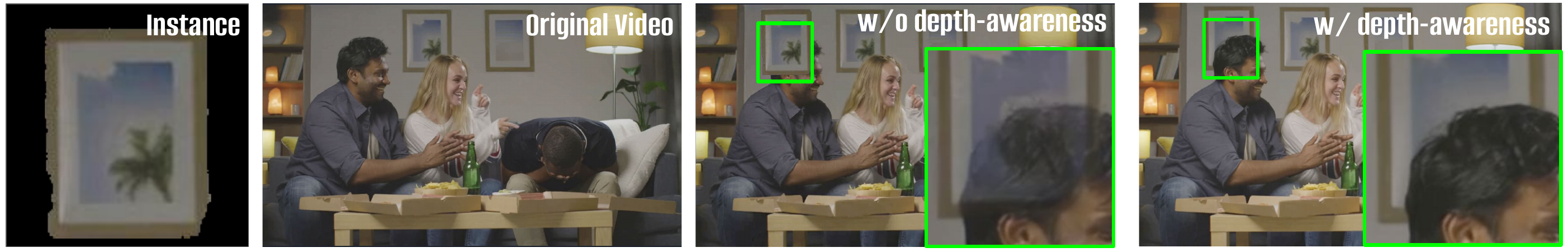}
\caption{\textbf{Visualization of the depth-aware insertion}. Conditioning on depth improves depth ordering and occlusion handling, reducing foreground/background blending artifacts compared to a depth-agnostic variant.}
\label{fig:depth_awareness}
\vspace{-1em}
\end{figure*}

\paragraph{Variable-Information Guidance (VIG).} 
Precise video instance insertion requires flexibility regarding user input, which can range from a single keyframe to dense per-frame annotations. To ensure robustness across varying levels of control, we introduce \textbf{Variable-Information Guidance (VIG)}, a dynamic contextual dropout strategy that samples an availability mask $A$ during training. This approach exposes the model to diverse supervision regimes, encouraging it to propagate instance information under sparse guidance while maintaining appearance fidelity and spatial alignment under dense guidance.

We employ a hybrid sampling strategy that covers the spectrum of annotation densities. This includes an extreme sparsity regime where only a single frame is retained, varying levels of sparse and dense sampling to simulate different user efforts, and an anchor regime that provides full-frame supervision. This strategy enables a smooth trade-off between temporal propagation and fidelity: under sparse conditions, the model learns to infer plausible motion by leveraging background context, while dense regimes prevent identity drift and preserve detailed appearance.

\subsection{Distribution-Preserving Temporal Masking (DPTM)}
\label{sec:dptm}

\begin{wrapfigure}{r}{0.5\columnwidth}
    \centering
    \includegraphics[width=\linewidth]{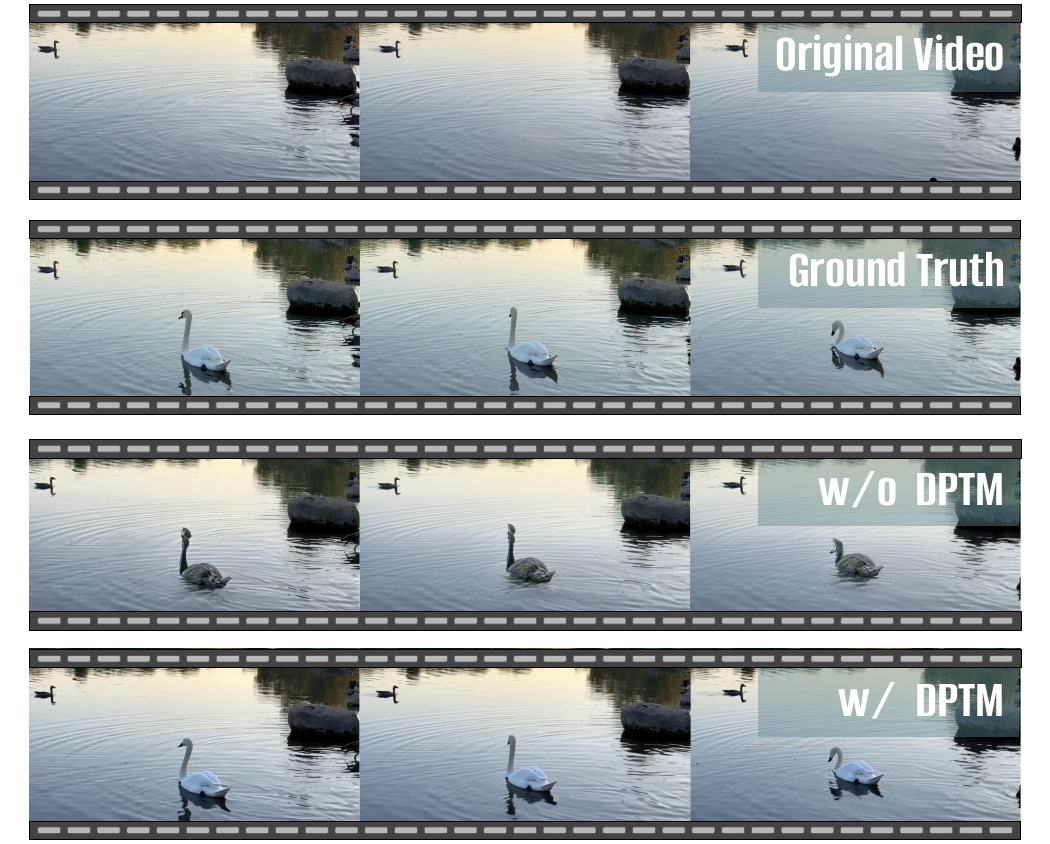}
    \caption{\textbf{Effectiveness of DPTM under sparse guidance.} 
    We compare results given segmented instance inputs only on odd frames. 
    Na\"ive masking leads to distribution shifts and temporal artifacts, whereas DPTM preserves encoder input statistics and significantly improves temporal stability.}
    \label{fig:DPTM}
\end{wrapfigure}

Modern video generative models~\cite{wan2025wan, chen2025goku, liu2025infinitystar} often rely on pretrained temporal video VAEs~\cite{yang2024cogvideox, wang2024omnitokenizer, zheng2024open, agarwal2025cosmos} that compress the temporal dimension by a factor $C_t$. If we na\"ively mask missing frames using $A$ in pixel space, the temporal VAE receives out-of-distribution inputs, which leads to flickering, miscoloring, and incomplete instance rendering as illustrated in \Cref{fig:DPTM}. To address this, we propose \textbf{Distribution-Preserving Temporal Masking (DPTM)} to decouple distribution preservation from information masking, while staying fully consistent with the sampled availability mask $A$.

\paragraph{Pixel-space temporal completion.}
As shown in \Cref{fig:pipeline}, given a masked instance clip $I^A$, we first perform nearest interpolation in pixel space. Concretely, missing frames are filled by propagating the temporally nearest available instance frame forward and backward. This produces a temporally complete clip with natural video statistics, which helps keep the input distribution compatible with pretrained temporal VAEs.

\paragraph{Token-space masking.}
We then encode the interpolated clip into video tokens. Tokens corresponding to the originally unobserved frames are masked out in the latent space. This explicit masking enables the model to distinguish valid signals from interpolated fillers, thereby accelerating convergence.

\paragraph{Availability channel aligned to token resolution.}
To distinguish observed frames from interpolated ones, we introduce an availability channel. 
Let $T'=\lceil T/C_t \rceil$ be the compressed token length and $(H',W')$ be the spatial resolution. 
We reshape the binary frame-level availability mask $A \in \{0,1\}^{T\times H \times W}$ into:
\[
A \in \mathbb{R}^{C_t \times T' \times H' \times W'}.
\]
This transformation moves the local temporal window $C_t$ into the channel dimension, preserving fine-grained observability patterns within each compressed token. 
The resulting tensor is concatenated with other conditions and processed by the VACE context adapter. 
Combined with pixel-space completion and token-space masking, this strategy enables precise sparse conditioning and minimizes temporal artifacts, as evidenced in \Cref{fig:DPTM}.

\subsection{Geometry- and Appearance-Robust Training}
\label{sec:robustness_training}

\begin{wrapfigure}{r}{0.5\columnwidth}
    \centering
    \includegraphics[width=\linewidth]{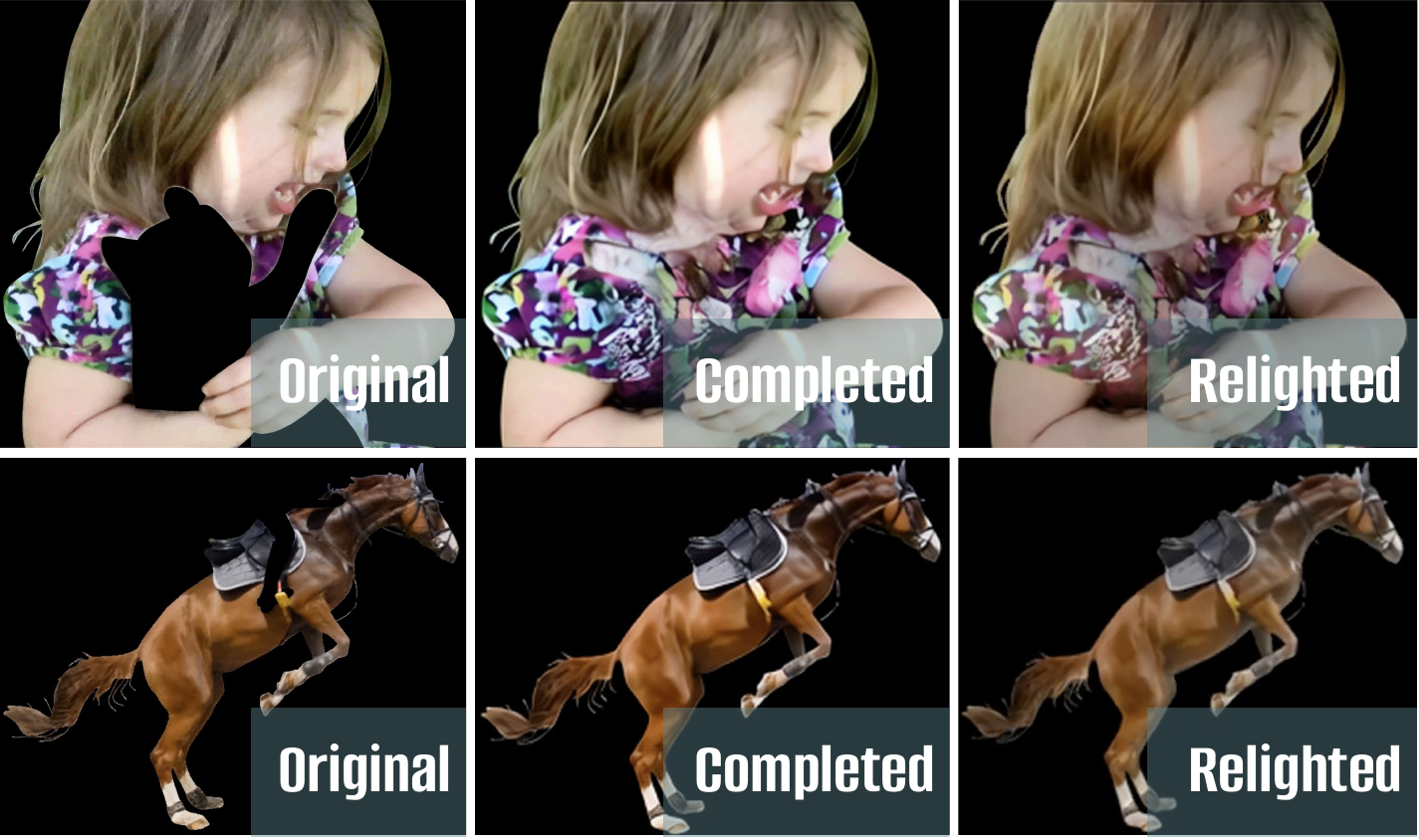}
    \caption{\textbf{Visualization of amodal instance completion and relighting augmentations.} We complete occluded instance cutouts to form pseudo-amodal inputs and apply moderate relighting to improve illumination compatibility with the target scene while preserving keyframe appearance controllability.}
    \vspace{-4mm}
    \label{fig:CNR}
\end{wrapfigure}

We further incorporate three complementary strategies to improve geometric plausibility and appearance robustness: depth-aware conditioning, occlusion-aware completion augmentation, and relighting augmentation.

\paragraph{Depth-aware insertion.}
Purely 2D video editing lacks explicit geometric reasoning, making scale, depth ordering, and occlusion ambiguous. We therefore condition the model on two depth signals: the background depth $D_V$ computed from $V$, and the instance depth $D_I$ extracted from $\hat{V}$ using mask $M$ and masked by availability as $D_I^A$. These depth cues are provided alongside RGB, spatial mask, and availability signals, enabling the model to reason about relative depth ordering and produce physically plausible layer compositing. As shown in \Cref{fig:depth_awareness}, Depth conditioning helps the model respect depth ordering and occlusion relations during insertion.


\begin{wrapfigure}{r}{0.5\columnwidth}
    \centering
    \vspace{-4mm}
    \includegraphics[width=\linewidth]{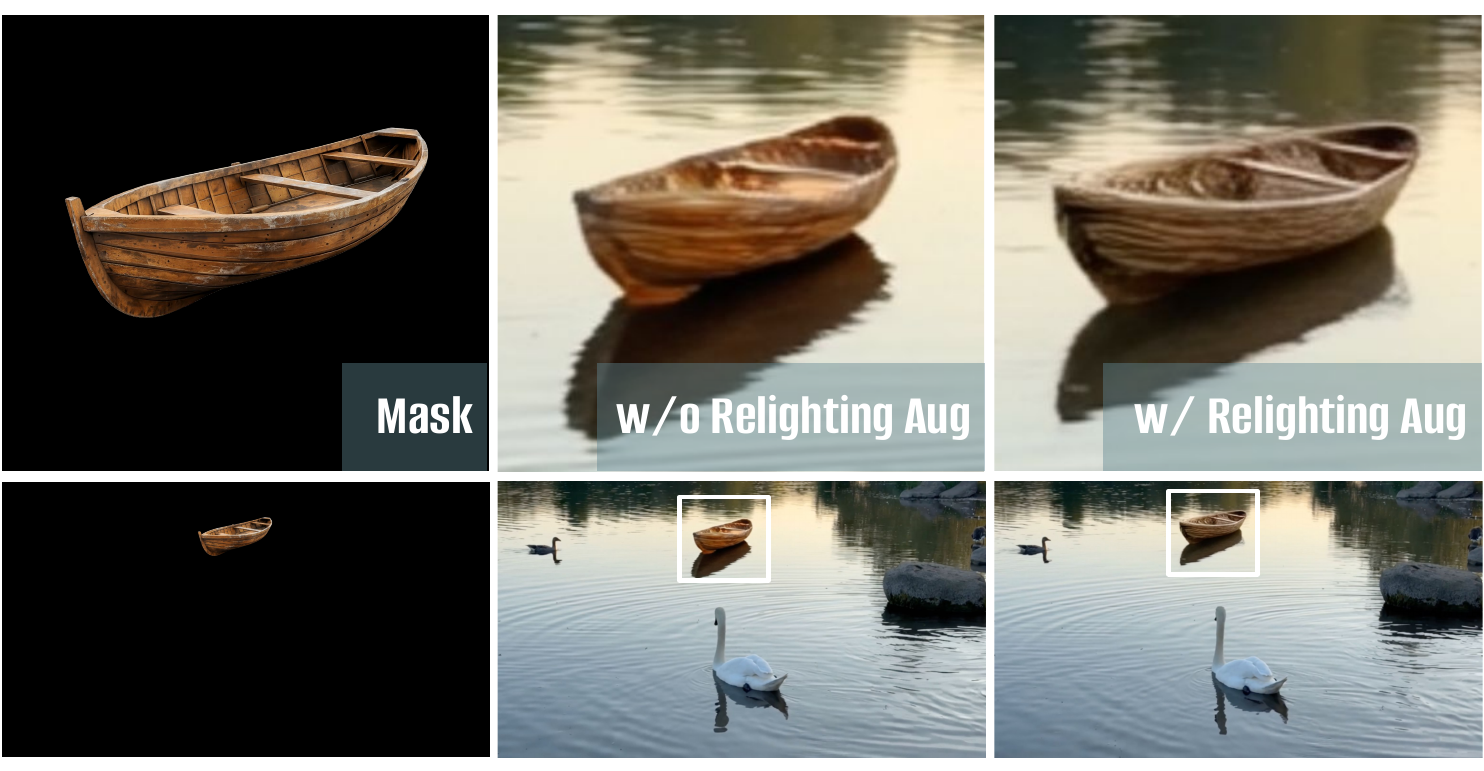}
    \caption{\textbf{Effectiveness of relighting augmentation.} Without relighting, the model over-preserves the input instance appearance, causing illumination mismatch after insertion. Relighting augmentation encourages lighting adaptation, yielding more coherent blending with the surrounding scene.}
    \label{fig:relight_aug}
\end{wrapfigure}

\paragraph{Amodal instance augmentation.}
During inference, users typically provide spatially complete (amodal~\cite{ao2023image, ao2025open}) instances, expecting the model to resolve occlusions based on scene geometry. 
However, training instances extracted directly from segmentation are often fragmented by scene occlusions. 
To align training with this inference requirement, we introduce an amodal augmentation strategy. 
Specifically, we reconstruct missing regions to generate a complete pseudo-amodal instance as the input condition, while supervising the model with the original occluded video. 
This pairing forces the model to learn the compositing logic required to correctly occlude a complete input instance according to depth cues. 
Details of the completion construction are provided in Sec.~\ref{sec:impl_data}. Some examples of amodal completion results are shown in \Cref{fig:CNR}.

\paragraph{Instance relighting augmentation.}
In video instance insertion, strictly preserving the provided instance appearance can lead to incompatible lighting between the instance and the background scene. To improve automatic lighting adaptation, we augment training with relighted instances generated by IC-Light~\cite{zhang2025scaling}. For each instance clip, we synthesize a relighted version $\tilde{I}$ under randomly sampled background lighting conditions. The relighted instance $\tilde{I}$ is then used as additional training data for instance conditioning. \Cref{fig:relight_aug} shows the qualitative results of the relighting augmentation.

\section{Implementation Details}
\label{sec:implementation_details}

\subsection{Data Preparation}
\label{sec:impl_data}

We construct training tuples of the form $(\ \hat{V},\ V,\ I,\ M,\ D_V,\ D_I\ )$,
where $\hat{V}$ is the target video containing the instance, $V$ is the paired background video, $(I,M)$ are the segmented instance clip and its mask, and $(D_V,D_I)$ are depth maps computed for background and instance signals.

\paragraph{Data sources.}
We use real-world videos from ROSE~\cite{miao2025rose}, VPData~\cite{bian2025videopainter}, MOSE~\cite{MOSE, MOSEv2}, and DAVIS~\cite{Caelles_arXiv_2019, Pont-Tuset_arXiv_2017}. We filter the segmentation results to favor clips with clean and logically removable instances and discard cases with ambiguous or non-detachable masks. After filtering, we obtain 16{,}642 video clips with corresponding foreground segmentation masks, each with at least 49 frames.

\paragraph{Background video generation.}
To construct paired videos $\{\hat{V},V\}$, we train a side-effect-aware video instance removal model on ROSE~\cite{miao2025rose}. Given a target video $\hat{V}$ and an instance mask $M$, we extract the masked instance $I$ from $\hat{V}$ and apply the instance removal model to obtain the paired background video $V$. This procedure produces training pairs that share the same scene dynamics while differing in the presence of the target instance, which matches the insertion objective.

\paragraph{Depth estimation.}
We compute $D_V$ by running Depth Anything V3~\cite{lin2025depth} on the background video $V$. We compute $D_I$ by running it on the target video $\hat{V}$ and then cropping the instance region using $M$.



\paragraph{Amodal Completion Data Construction.}
To support amodal instance augmentation, we develop a completion model $\mathcal{F}$ that reconstructs the full amodal appearance $\tilde{I} = \mathcal{F}(I)$ from an occluded modal segment $I$. 
We construct paired training data by curating a collection of fully visible instances from our dataset to serve as amodal ground truth. Corresponding modal inputs are synthesized by randomly overlaying additional instance cutouts onto these samples, creating a mapping from occluded views to complete appearances. 
We employ a pre-trained image editing framework as the base generator, fine-tuning it via LoRA to specialize in amodal recovery. This model is applied frame-wise to generate pseudo-amodal instance clips. Finally, the amodal masks $\tilde{M}$ are obtained through thresholding, while the amodal depth $\tilde{D}_I$ is synthesized using interpolation from available depth values.



\subsection{Training Schedule}
\label{sec:training_details}

\paragraph{Training Hyperparameters.} Our models are built upon two pretrained backbones: \texttt{Wan2.1-VACE-1.3B} and \texttt{Wan2.2-VACE-Fun-A14B}, denoted as \OMs and \OMl, respectively. To ensure stable adaptation to our multi-condition setting while preserving pretrained generative priors, we adopt a staged training strategy with progressively increased model flexibility. 
All stages prior to Stage~V are conducted at a spatial resolution of $832\times480$ and temporal length of 49 frames to stabilize optimization and reduce computational cost.

\begin{itemize}[leftmargin=*, nosep]
    \item \textbf{Stage I: Adapter input warm-up.}  
    We replace the input projection of the VACE context adapter to support multi-channel conditioning and train only this newly introduced layer while freezing all other modules.  
    This warm-up stage stabilizes condition injection before finetuning deeper components.

    \item \textbf{Stage II: Adapter finetuning.}  
    We finetune the full VACE context adapter while keeping the diffusion backbone frozen.  
    This allows the adapter to better align pretrained representations with task-specific conditioning without disturbing the backbone priors.

    \item \textbf{Stage III: Joint finetuning.}  
    We jointly finetune the context adapter and the diffusion backbone.  
    This step improves the coupling between conditioning signals and generation dynamics.

    \item \textbf{Stage IV: Augmented training.}  
    We introduce occlusion-aware completion and relighting augmentations and continue joint training using parameter-efficient LoRA finetuning.  
    This enhances robustness under challenging visibility and appearance variations.

    \item \textbf{Stage V: Resolution and temporal extension.}  
    We further extend the model from generating 49-frame videos at $832\times480$ resolution to 120-frame videos at $1280\times720$ resolution.  
\end{itemize}

All experiments are trained using AdamW on NVIDIA H100 GPUs. Wan2.2 adopts a dual-denoiser architecture with separate high-noise and low-noise denoising stages; therefore, for \OMl, the two denoisers are trained independently using the same schedule and hyperparameters.

\section{Experiments}

\subsection{Evaluation Data.}

We introduce \OMb, a curated evaluation benchmark constructed on top of the BURST dataset~\cite{athar2023burst}. \OM-Bench is designed to assess instance-level video editing under diverse real-world conditions, including urban driving scenes, scripted movie clips, and in-the-wild internet videos.

Starting from BURST, we carefully select 100 videos with broad scenario coverage and no overlap with our training data. For each video, we manually inspect the instance segmentation annotations and correct missing or low-quality masks to ensure reliable conditioning inputs. To obtain clean background videos, we remove the target instances using ROSE~\cite{miao2025rose}, a side-effect-aware instance removal model. 
Throughout this section, in the context of video instance insertion, we refer to these processed background videos as \say{clean videos} and the videos with instances as \say{target videos}.

\subsection{Evaluation Protocol.}
We report results for both \OMs and \OMl under two sparse-control settings:
(i) \textbf{First-frame control}, where the segmented instance image and mask are provided only at the first frame;
(ii) \textbf{First \& last-frame control}, where the segmented instance image and mask are provided at both the first and last frames.
For all experiments, we perform inference using 50 diffusion denoising steps.

\subsection{Baselines.}
Our goal is precise video instance insertion under sparse user control, where the user provides only a small number of segmented instance frames to specify the appearance and placement of the instance. Since there are no directly comparable open-sourced models that support the same sparse conditioning interface, we compare against representative approaches that can achieve similar functionality via different pipelines. Specifically, we consider three categories:

\textbf{(1) Agentic pipeline: image editing + image-to-video (I2V) generation.}
This pipeline first edits the first frame (and optionally the last frame) using the reference instance and mask, and then generates the full video conditioned on the edited frame(s) with an I2V model. This is a commonly adopted recipe in recent editing data generation workflows~\cite{he2025openve, chen2025ivebench}. In our experiments, we use \texttt{Nano-banana-Pro}~\cite{GoogleCloud2025BananaPro} for image editing, taking the unedited frame(s) together with the segmented reference instance image and mask as input. We then use \texttt{Wan2.2-Fun-A14B-InP}~\cite{wan2025wan} for I2V generation under both first-frame and first \& last-frame control.

\begin{wraptable}{r}{0.5\textwidth}
\centering
\scriptsize
\setlength{\tabcolsep}{3pt}
\vspace{-4mm}
\caption{\textbf{Input conditions for all compared methods.} We indicate whether each approach uses the clean video, reference instance frames (first/last), and instance masks (sparse or dense), as well as an optional text prompt. Notably, several baselines require dense per-frame masks, whereas PISCO operates with sparse keyframe masks (first-only or first\&last).}
\label{tab:method_inputs}
\resizebox{\linewidth}{!}{%
\begin{tabular}{lccccccc}
\toprule
\multirow{2}{*}{Method} & & \multicolumn{2}{c}{Instance} & \multicolumn{3}{c}{Instance Mask} & \\
 \cmidrule(lr){3-4} \cmidrule(lr){5-7} 
& Video & First & Last & First & Last & All & Text \\
\midrule
\multicolumn{8}{c}{\textit{Image Editing + I2V Generation}} \\
\midrule
First & \cmark & \cmark &  & \cmark &  &  & \cmark \\
First\&Last & \cmark & \cmark & \cmark & \cmark & \cmark &  & \cmark \\
\midrule
\multicolumn{8}{c}{\textit{Video Inpainting}} \\
\midrule
CoCoCo & \cmark &  &  & \cmark & \cmark & \cmark & \cmark \\
VideoPainter & \cmark &  &  & \cmark & \cmark & \cmark & \cmark \\
\midrule
\multicolumn{8}{c}{\textit{Video-to-Video Editing}} \\
\midrule
VACE (14B) & \cmark & \cmark &  & \cmark & \cmark & \cmark & \cmark \\
UniVideo & \cmark & \cmark &  &  &  &  & \cmark \\
\midrule
\multicolumn{8}{c}{\textit{\OM 1.3B (Ours)}} \\
\midrule
First & \cmark & \cmark &  & \cmark &  &  &  \\
First\&Last & \cmark & \cmark & \cmark & \cmark & \cmark &  &  \\
\midrule
\multicolumn{8}{c}{\textit{\OM 14B (Ours)}} \\
\midrule
First & \cmark & \cmark &  & \cmark &  &  &  \\
First\&Last & \cmark & \cmark & \cmark & \cmark & \cmark &  &  \\
\bottomrule
\end{tabular}%
}
\vspace{-4mm}
\end{wraptable}

\textbf{(2) Video inpainting models.}
Video inpainting methods typically require dense masks over all frames, along with a text prompt describing the inserted object and the target scene. In this experiment, we use \texttt{Qwen3-VL-32B-Instruct} to generate a detailed textual description of the segmented reference instance image as the prompt. We adopt CoCoCo~\cite{zi2025cococo} and VideoPainter~\cite{bian2025videopainter} as strong inpainting baselines. Both take the text prompt and the masked video as input, and this setting is also widely used in video editing pipelines~\cite{he2025openve, chen2025ivebench}.

\textbf{(3) Reference-guided video-to-video editing models.}
Another alternative is generic video-to-video editing, which edits the entire video in an end-to-end manner using multiple conditions such as a reference image, text prompt, and masks. In this category, we evaluate VACE~\cite{jiang2025vace} and UniVideo~\cite{wei2025univideo}. VACE takes the unedited video, segmented reference image, text prompt, and the full mask video as input. UniVideo does not support mask conditioning; it uses only the unedited video, segmented reference image, and text prompt.

\Cref{tab:method_inputs} summarizes the input conditions for all methods. we note that several baselines requires dense per-frame masks (the full mask video), whereas our models work with sparse masks (first frame or first \& last frame). This comparison reflects the design philosophy of \OM: minimizing user effort while maintaining controllability and output quality. Consider that different pipelines support different video resolutions and frame length, we keep the experimental comparison with 832 pixel width, 480 pixel height, and 49 frames.

\subsection{Quantitative Experiments.}

\paragraph{Reference-based Video Quality Assessment.}

For the Precise Video Instance Insertion task, we use the \OMb dataset's ground-truth pairs $(V, \hat{V})$ to conduct reference-based evaluation. The generated videos $V'$ are assessed using four standard metrics: FVD~\cite{skorokhodov2022stylegan}, LPIPS~\cite{zhang2018unreasonable}, PSNR~\cite{jahne2005digital}, and SSIM~\cite{wang2004image}. To ensure a comprehensive analysis, we evaluate at two levels:
(1) \textbf{Whole-video assessment} calculates metrics between $V'$ and $\hat{V}$ to measure global consistency and the accuracy of physical effects like shadows or reflections. 
(2) \textbf{Foreground assessment} isolates the inserted instance using mask $M$, computing $\text{Metric}(V' \odot M, \hat{V} \odot M)$. This directly verifies whether the generated object adheres to the reference conditions and intended trajectory.

\begin{table*}[t]
\centering
\scriptsize
\setlength{\tabcolsep}{10pt}
\vspace{-2mm}
\caption{\textbf{Quantitative Results (FVD, LPIPS, PSNR, SSIM).} Comparisons on Whole Video and Foreground Region. \textcolor{blue}{Blue} and \textcolor{red}{red} indicate best and second-best among standard settings. \textcolor{gray}{Gray rows} indicate the Five-Frame reference setting (not included in ranking).}
\label{tab:quantitative_metrics}
\renewcommand{\arraystretch}{1}
\resizebox{\textwidth}{!}{%
\begin{tabular}{lcccc|cccc}
\toprule
\multirow{2}{*}{Method} & \multicolumn{4}{c|}{Whole Video} & \multicolumn{4}{c}{Foreground} \\
\cmidrule(lr){2-5}
\cmidrule(lr){6-9}
 & FVD \colordown & LPIPS \colordown & PSNR \colorup & SSIM \colorup & FVD \colordown & LPIPS \colordown & PSNR \colorup & SSIM \colorup \\
\midrule
\multicolumn{9}{c}{\textit{Image Editing + I2V Generation}} \\
\midrule
First Only & 826 & 0.451 & 15.47 & 0.55 & 297 & 0.030 & 30.24 & 0.97 \\
First + Last & 624 & 0.392 & 16.44 & 0.56 & 250 & 0.030 & 30.38 & \cellcolor{blue!15}0.98 \\
\midrule
\multicolumn{9}{c}{\textit{Video Inpainting}} \\
\midrule
CoCoCo & 590 & 0.191 & 23.62 & 0.80 & 398 & 0.031 & 30.26 & 0.97 \\
VideoPainter & 524 & 0.154 & 23.11 & 0.78 & 384 & 0.035 & 29.27 & 0.97 \\
\midrule
\multicolumn{9}{c}{\textit{Video-to-Video Editing}} \\
\midrule
VACE$_{\text{14B}}$ & 371 & \cellcolor{red!15}0.103 & 25.55 & \cellcolor{red!15}0.88 & 273 & 0.028 & 30.55 & \cellcolor{blue!15}0.98 \\
UniVideo & 485 & 0.211 & 19.22 & 0.61 & 310 & 0.031 & 29.21 & 0.97 \\
\midrule
\multicolumn{9}{c}{\textit{\OM$_{\text{1.3B}}$ (Ours)}} \\
\midrule
First Only & 398 & 0.121 & 25.35 & 0.87 & 243 & 0.029 & 30.51 & \cellcolor{blue!15}0.98 \\
First + Last & \cellcolor{red!15}269 & \cellcolor{red!15}0.103 & \cellcolor{blue!15}27.01 & \cellcolor{red!15}0.88 & \cellcolor{red!15}171 & \cellcolor{blue!15}0.024 & \cellcolor{red!15}32.99 & \cellcolor{blue!15}0.98 \\
\color{gray}Five Frames & \color{gray}172 & \color{gray}0.089 & \color{gray}28.53 & \color{gray}0.89 & \color{gray}104 & \color{gray}0.018 & \color{gray}35.07 & \color{gray}0.98 \\
\midrule
\multicolumn{9}{c}{\textit{\OM$_{\text{14B}}$ (Ours)}} \\
\midrule
First Only & 337 & 0.116 & 24.81 & \cellcolor{red!15}0.88 & 222 & 0.029 & 30.61 & 0.97 \\
First + Last & \cellcolor{blue!15}204 & \cellcolor{blue!15}0.097 & \cellcolor{red!15}26.58 & \cellcolor{blue!15}0.89 & \cellcolor{blue!15}138 & \cellcolor{red!15}0.022 & \cellcolor{blue!15}33.58 & \cellcolor{blue!15}0.98 \\
\color{gray}Five Frames & \color{gray}136 & \color{gray}0.084 & \color{gray}28.01 & \color{gray}0.90 & \color{gray}75 & \color{gray}0.015 & \color{gray}35.94 & \color{gray}0.98 \\
\bottomrule
\end{tabular}%
}
\vspace{-4mm}
\end{table*}

Quantitative results in \Cref{tab:quantitative_metrics} show that \OM consistently outperforms all baselines. In whole-video metrics, the \textit{Image Editing + I2V} pipeline performs worst due to severe background hallucination, as reflected by its low PSNR and SSIM. While \textit{Video Inpainting} models like CoCoCo and VideoPainter preserve unmasked regions better, their high FVD scores indicate poor temporal coherence. VACE emerges as the strongest baseline, yet \OM-14B (First \& Last) surpasses it significantly, reducing FVD from 371 to 204 and LPIPS from 0.103 to 0.097. In foreground assessment, \OM-14B (First \& Last) achieves a Foreground FVD of 138 and LPIPS of 0.022, both substantially better than competing methods. These results demonstrate that our framework generates instances with high visual quality and strict temporal alignment. Moreover, the superior performance of the First \& Last configuration over First Only confirms that endpoint constraints effectively stabilize complex instance dynamics.

We also evaluated scalability using a ``Five Frames'' setting, incorporating the first, last, and three random intermediate frames. Unlike existing baselines that lack the flexibility for arbitrary multi-frame conditioning, \OM leverages these additional signals to further refine quality. For example, \OM-14B with five frames improves whole-video FVD to 136 and Foreground LPIPS to 0.015. This performance boost underscores our framework's unique ability to effectively scale controllability with additional sparse inputs.

\begin{table*}[ht]
\centering
\scriptsize
\setlength{\tabcolsep}{5pt}
\vspace{-1mm}
\caption{\textbf{VBench Quantitative Comparison.} We compare \OM against recent baselines. To ensure precise evaluation for instance insertion, \textit{Background} and \textit{Subject Consistency} are computed using masked regions to isolate foreground/background signals. \textcolor{blue}{Blue} and \textcolor{red}{red} indicate the best and second-best results among standard settings. \textcolor{gray}{Gray rows} indicate the Five-Frame setting.}
\label{tab:vbench_evaluation}
\renewcommand{\arraystretch}{1}
\resizebox{\textwidth}{!}{%
\begin{tabular}{lccccccccc}
\toprule
Method & \shortstack{Background\\Consistency} & \shortstack{Subject\\Consistency} & \shortstack{Aesthetic\\Quality} & \shortstack{Imaging\\Quality} & \shortstack{Motion\\Smoothness} & \shortstack{Overall\\Consistency} & \shortstack{Temporal\\Flickering} & \shortstack{Temporal\\Style} & Average \\
\midrule
\multicolumn{10}{c}{\textit{Image Editing + I2V Generation}} \\
\midrule
First Only & 91.43 & 83.86 & 48.83 & 59.01 & 97.89 & 15.11 & 95.89 & 15.11 & 63.39 \\
First + Last & 92.28 & 85.09 & 49.47 & 59.71 & 98.18 & \cellcolor{red!15}15.58 & 96.26 & 15.58 & 64.02 \\
\midrule
\multicolumn{10}{c}{\textit{Video Inpainting}} \\
\midrule
CoCoCo & \cellcolor{blue!15} 94.63 & 89.55 & 47.81 & 55.76 & 98.67 & 14.61 & \cellcolor{red!15} 97.69 & 14.61 & 64.17 \\
VideoPainter & \cellcolor{red!15} 94.51 & 89.14 & 47.82 & 57.68 & \cellcolor{blue!15} 98.97 & 13.85 & \cellcolor{blue!15} 97.77 & 13.85 & 64.20 \\
\midrule
\multicolumn{10}{c}{\textit{Video-to-Video Editing}} \\
\midrule
VACE$_{\text{14B}}$ & 94.21 & 90.29 & 48.69 & 60.95 & \cellcolor{red!15} 98.90 & 14.87 & 97.56 & 14.87 & 65.04 \\
UniVideo & 94.04 & 89.88 & 49.41 & 60.84 & 98.80 & 15.51 & 97.01 & \cellcolor{blue!15} 15.92 & 65.18 \\
\midrule
\multicolumn{10}{c}{\textit{\OM$_{\text{1.3B}}$ (Ours)}} \\
\midrule
First Only & 93.72 & 87.16 & 47.70 & 60.67 & 98.85 & 14.92 & 97.54 & 14.92 & 64.43 \\
First + Last & 94.07 & \cellcolor{red!15} 91.33 & \cellcolor{red!15} 49.48 & \cellcolor{red!15} 61.18 & 98.86 & \cellcolor{red!15} 15.58 & 97.51 & 15.58 & \cellcolor{red!15} 65.45 \\
\color{gray}Five Frames & \color{gray}94.23 & \color{gray}91.45 & \color{gray}50.01 & \color{gray}62.09 & \color{gray}98.86 & \color{gray}15.61 & \color{gray}97.67 & \color{gray}15.61 & \color{gray}65.89 \\
\midrule
\multicolumn{10}{c}{\textit{\OM$_{\text{14B}}$ (Ours)}} \\
\midrule
First Only & 93.84 & 87.26 & 48.24 & 60.80 & 98.79 & 15.14 & 97.26 & 15.14 & 64.56 \\
First + Last & 94.20 & \cellcolor{blue!15} 91.57 & \cellcolor{blue!15} 50.08 & \cellcolor{blue!15} 62.00 & 98.79 & \cellcolor{blue!15} 15.64 & 97.21 & \cellcolor{red!15} 15.64 & \cellcolor{blue!15} 65.64 \\
\color{gray}Five Frames & \color{gray}94.42 & \color{gray}91.98 & \color{gray}51.45 & \color{gray}62.87 & \color{gray}98.89 & \color{gray}15.82 & \color{gray}97.34 & \color{gray}15.57 & \color{gray}66.04 \\
\bottomrule
\end{tabular}%
}
\vspace{-2mm}
\end{table*}

\paragraph{Reference-free Video Quality Assessment.}
To evaluate perceptual quality and temporal consistency without pixel-aligned ground truth~\cite{he2024cover}, we employ VBench~\cite{huang2024vbench}. However, standard VBench consistency metrics compute feature similarity over entire frames, mixing foreground and background signals. This is suboptimal for instance insertion, where object consistency and background preservation should be assessed independently. Therefore, we utilize instance masks to isolate foreground and background regions, extracting CLIP/DINO features solely from these areas to compute consistency scores. Other metrics, such as Aesthetic Quality and Motion Smoothness, follow the official implementation.

The results in \Cref{tab:vbench_evaluation} show that \OM demonstrates superior performance across most perceptual dimensions. Notably, in Subject Consistency, \OM-14B (First \& Last) achieves a score of 91.57, significantly outperforming VACE (90.29) and inpainting models ($\sim$89). This indicates that our model maintains the identity and appearance of the inserted instance more faithfully over time. \OM also secures top spots in Imaging Quality and Aesthetic Quality, suffering less from artifacts compared to baseline pipelines. While inpainting baselines naturally achieve high Background Consistency by copying pixels, \OM remains highly competitive (94.20) while offering superior motion and instance fidelity.

Following the reference-based evaluation, we also report \OM's performance using the ``Five Frames'' setting. Adding three intermediate control frames further boosts metrics, with \OM-14B reaching 91.98 in Subject Consistency and 51.45 in Aesthetic Quality. This consistent upward trend across both reference-based and reference-free benchmarks reinforces our conclusion: \OM is uniquely capable of leveraging additional sparse visual prompts to refine generation quality, a flexibility not supported by current state-of-the-art baselines.

\begin{figure*}[!t]
\centering
\includegraphics[width=\textwidth]{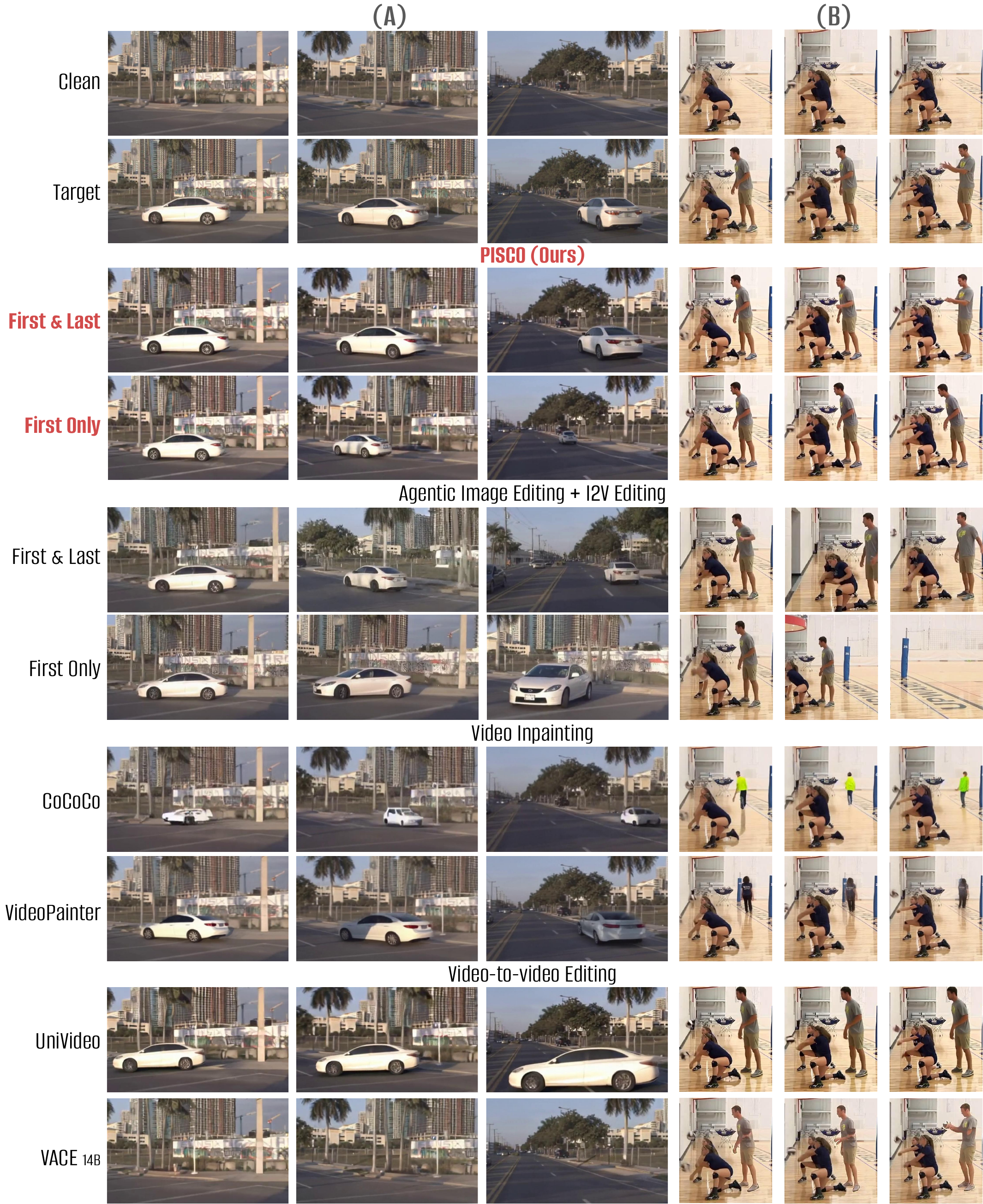}
\vspace{1mm}
\caption{\textbf{Qualitative Comparison}. Visual results of instance insertion in two dynamic scenes (A) and (B). Existing methods struggle with background preservation (Agentic), instance identity (Inpainting), or scale consistency (UniVideo/VACE). Our method, \OM, achieves superior visual fidelity and spatiotemporal alignment, particularly when utilizing sparse "First \& Last" frame control.}

\label{fig:qualitative}
\vspace{-3mm}
\end{figure*}


\subsection{Qualitative Experiments}

We qualitatively compare generated frames in \Cref{fig:qualitative} to assess different methods. The results highlight several baseline limitations: agentic pipelines (combining image editing with I2V) often regenerate the entire sequence, severely degrading background consistency. Inpainting models like CoCoCo and VideoPainter, lacking reference instance inputs, frequently hallucinate blurry or incorrect objects. Meanwhile, reference-based V2V models such as UniVideo and VACE struggle with long-term spatial control; they often fail to maintain correct perspective (e.g., the oversized car in UniVideo) or lose the target instance entirely in later frames. 

In contrast, our \OM approach demonstrates superior controllability. While first-frame control alone effectively captures appearance, the trajectory may slightly drift over time. However, as shown in the ``\OM First \& Last'' results, providing sparse control frames easily resolves this ambiguity, ensuring precise alignment with the target trajectory throughout the sequence.


\section{Beyond Instance Insertion: Broader Applications of PISCO}
\label{sec:beyond_insertion}

\begin{figure*}[t]
    \centering
    \includegraphics[width=\textwidth]{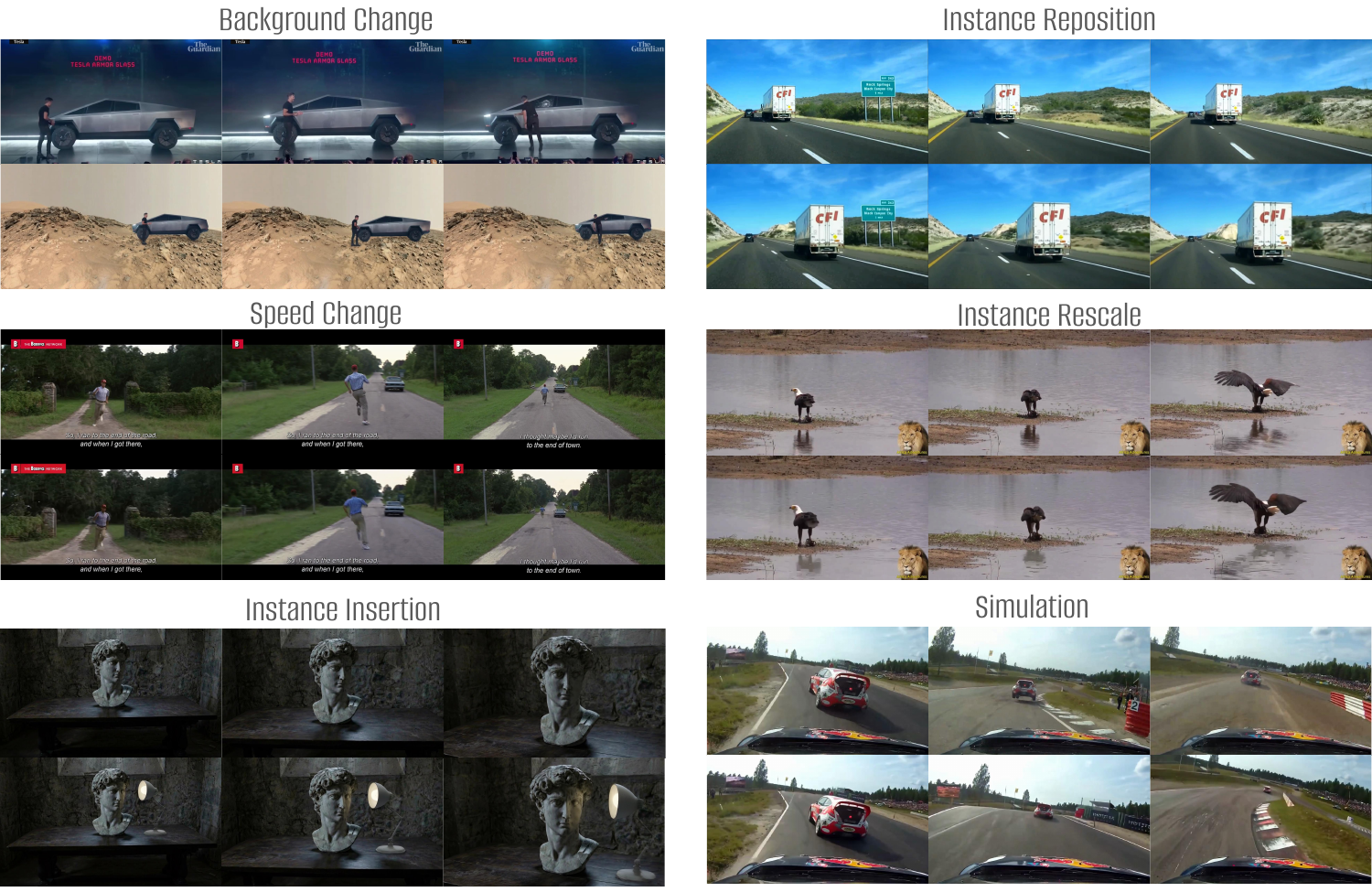}
    \caption{\textbf{Broader applications beyond instance insertion.}
    Leveraging the same instance-level conditioning and temporal propagation, \OM supports diverse instance-centric video edits and simulation
    under controllable spatiotemporal guidance: \textit{Background Change}, \textit{Instance Repositioning}, \textit{Speed Change}, \textit{Instance Rescaling},
    and \textit{Dynamics Simulation}.}
    \label{fig:broader_apps}
    \vspace{-1em}
\end{figure*}

While \OM is designed for precise video instance insertion under sparse keyframe control, the same
instance-level conditioning and temporal propagation machinery naturally generalizes to a broader set
of controllable video editing and simulation tasks.
Figure~\ref{fig:broader_apps} illustrates representative use cases enabled by \OM.
In particular, \OM can (i) perform \textbf{background change} by re-rendering the surrounding scene while
preserving the foreground instance identity and motion; (ii) support \textbf{instance repositioning} and
\textbf{instance rescaling} by adjusting the instance location and size while maintaining scene-consistent
interactions (e.g., occlusions and shadows); (iii) realize \textbf{speed change}~\cite{gao2026pulsemotion} by temporally subsampling
instance-related conditions to induce faster or slower motion; and (iv) enable \textbf{dynamics simulation}
by providing partial instance-related conditions to create counterfactual trajectories for stress-testing
downstream perception and planning systems~\cite{wang2026generativeaiad}.
These extensions highlight \OM as a general-purpose, instance-centric video editing framework rather than a
single-task solution.
We encourage future researchers and developers to further investigate these directions and extend \OM to
additional instance-centric editing and simulation scenarios.

\section{Conclusion}

In this paper, we presented \OM, a video diffusion model for precise video instance insertion under sparse user control. By introducing Variable-Information Guidance and Distribution-Preserving Temporal Masking, together with geometry-aware conditioning, \OM effectively resolves the distribution shift and temporal instability issues that arise when applying sparse conditions to pretrained video diffusion models. Extensive experiments on the proposed \OM-Bench demonstrate that \OM consistently outperforms strong inpainting, video editing, and agentic baselines, while exhibiting clear and monotonic performance gains as additional sparse control frames are provided. These results position \OM as a practical and scalable solution for professional-grade video editing, and a key step toward highly controllable, low-effort, AI-assisted filmmaking.




\newpage
{
\small
\bibliographystyle{IEEEtran}
\bibliography{main}

@String(ICCV  = {Int. Conf. Comput. Vis.})

@String(AAAI  = {AAAI})

@String(ICCV  = {ICCV})

@inproceedings{chen2025goku,
  title={Goku: Flow based video generative foundation models},
  author={Chen, Shoufa and Ge, Chongjian and Zhang, Yuqi and Zhang, Yida and Zhu, Fengda and Yang, Hao and Hao, Hongxiang and Wu, Hui and Lai, Zhichao and Hu, Yifei and others},
  booktitle={Proceedings of the Computer Vision and Pattern Recognition Conference},
  pages={23516--23527},
  year={2025}
}

@article{liu2025infinitystar,
  title={Infinitystar: Unified spacetime autoregressive modeling for visual generation},
  author={Liu, Jinlai and Han, Jian and Yan, Bin and Wu, Hui and Zhu, Fengda and Wang, Xing and Jiang, Yi and Peng, Bingyue and Yuan, Zehuan},
  journal={arXiv preprint arXiv:2511.04675},
  year={2025}
}

@article{yang2024cogvideox,
  title={Cogvideox: Text-to-video diffusion models with an expert transformer},
  author={Yang, Zhuoyi and Teng, Jiayan and Zheng, Wendi and Ding, Ming and Huang, Shiyu and Xu, Jiazheng and Yang, Yuanming and Hong, Wenyi and Zhang, Xiaohan and Feng, Guanyu and others},
  journal={arXiv preprint arXiv:2408.06072},
  year={2024}
}

@article{zheng2024open,
  title={Open-sora: Democratizing efficient video production for all},
  author={Zheng, Zangwei and Peng, Xiangyu and Yang, Tianji and Shen, Chenhui and Li, Shenggui and Liu, Hongxin and Zhou, Yukun and Li, Tianyi and You, Yang},
  journal={arXiv preprint arXiv:2412.20404},
  year={2024}
}

@article{agarwal2025cosmos,
  title={Cosmos world foundation model platform for physical ai},
  author={Agarwal, Niket and Ali, Arslan and Bala, Maciej and Balaji, Yogesh and Barker, Erik and Cai, Tiffany and Chattopadhyay, Prithvijit and Chen, Yongxin and Cui, Yin and Ding, Yifan and others},
  journal={arXiv preprint arXiv:2501.03575},
  year={2025}
}

@article{wang2024omnitokenizer,
  title={Omnitokenizer: A joint image-video tokenizer for visual generation},
  author={Wang, Junke and Jiang, Yi and Yuan, Zehuan and Peng, Bingyue and Wu, Zuxuan and Jiang, Yu-Gang},
  journal={Advances in Neural Information Processing Systems},
  volume={37},
  pages={28281--28295},
  year={2024}
}

@article{quan2024deep,
  title={Deep learning-based image and video inpainting: A survey},
  author={Quan, Weize and Chen, Jiaxi and Liu, Yanli and Yan, Dong-Ming and Wonka, Peter},
  journal={International Journal of Computer Vision},
  volume={132},
  number={7},
  pages={2367--2400},
  year={2024},
  publisher={Springer}
}

@article{yang2023diffusion,
  title={Diffusion models: A comprehensive survey of methods and applications},
  author={Yang, Ling and Zhang, Zhilong and Song, Yang and Hong, Shenda and Xu, Runsheng and Zhao, Yue and Zhang, Wentao and Cui, Bin and Yang, Ming-Hsuan},
  journal={ACM computing surveys},
  volume={56},
  number={4},
  pages={1--39},
  year={2023},
  publisher={ACM New York, NY, USA}
}

@article{wan2025wan,
  title={Wan: Open and advanced large-scale video generative models},
  author={Wan, Team and Wang, Ang and Ai, Baole and Wen, Bin and Mao, Chaojie and Xie, Chen-Wei and Chen, Di and Yu, Feiwu and Zhao, Haiming and Yang, Jianxiao and others},
  journal={arXiv preprint arXiv:2503.20314},
  year={2025}
}

@article{labs2025flux,
  title={FLUX. 1 Kontext: Flow Matching for In-Context Image Generation and Editing in Latent Space},
  author={Labs, Black Forest and Batifol, Stephen and Blattmann, Andreas and Boesel, Frederic and Consul, Saksham and Diagne, Cyril and Dockhorn, Tim and English, Jack and English, Zion and Esser, Patrick and others},
  journal={arXiv preprint arXiv:2506.15742},
  year={2025}
}

@inproceedings{chang2019free,
  title={Free-form video inpainting with 3d gated convolution and temporal patchgan},
  author={Chang, Ya-Liang and Liu, Zhe Yu and Lee, Kuan-Ying and Hsu, Winston},
  booktitle={Proceedings of the IEEE/CVF international conference on computer vision},
  pages={9066--9075},
  year={2019}
}

@inproceedings{hu2020proposal,
  title={Proposal-based video completion},
  author={Hu, Yuan-Ting and Wang, Heng and Ballas, Nicolas and Grauman, Kristen and Schwing, Alexander G},
  booktitle={European Conference on Computer Vision},
  pages={38--54},
  year={2020},
  organization={Springer}
}

@inproceedings{wang2019video,
  title={Video inpainting by jointly learning temporal structure and spatial details},
  author={Wang, Chuan and Huang, Haibin and Han, Xiaoguang and Wang, Jue},
  booktitle={Proceedings of the AAAI conference on artificial intelligence},
  volume={33},
  number={01},
  pages={5232--5239},
  year={2019}
}

@inproceedings{xu2019deep,
  title={Deep flow-guided video inpainting},
  author={Xu, Rui and Li, Xiaoxiao and Zhou, Bolei and Loy, Chen Change},
  booktitle={Proceedings of the IEEE/CVF conference on computer vision and pattern recognition},
  pages={3723--3732},
  year={2019}
}

@inproceedings{kim2019deep,
  title={Deep video inpainting},
  author={Kim, Dahun and Woo, Sanghyun and Lee, Joon-Young and Kweon, In So},
  booktitle={Proceedings of the IEEE/CVF conference on computer vision and pattern recognition},
  pages={5792--5801},
  year={2019}
}

@inproceedings{li2020short,
  title={Short-term and long-term context aggregation network for video inpainting},
  author={Li, Ang and Zhao, Shanshan and Ma, Xingjun and Gong, Mingming and Qi, Jianzhong and Zhang, Rui and Tao, Dacheng and Kotagiri, Ramamohanarao},
  booktitle={European Conference on Computer Vision},
  pages={728--743},
  year={2020},
  organization={Springer}
}

@inproceedings{gao2020flow,
  title={Flow-edge guided video completion},
  author={Gao, Chen and Saraf, Ayush and Huang, Jia-Bin and Kopf, Johannes},
  booktitle={European Conference on Computer Vision},
  pages={713--729},
  year={2020},
  organization={Springer}
}

@inproceedings{zou2021progressive,
  title={Progressive temporal feature alignment network for video inpainting},
  author={Zou, Xueyan and Yang, Linjie and Liu, Ding and Lee, Yong Jae},
  booktitle={Proceedings of the IEEE/CVF Conference on Computer Vision and Pattern Recognition},
  pages={16448--16457},
  year={2021}
}

@inproceedings{li2022towards,
  title={Towards an end-to-end framework for flow-guided video inpainting},
  author={Li, Zhen and Lu, Cheng-Ze and Qin, Jianhua and Guo, Chun-Le and Cheng, Ming-Ming},
  booktitle={Proceedings of the IEEE/CVF conference on computer vision and pattern recognition},
  pages={17562--17571},
  year={2022}
}

@inproceedings{liu2021fuseformer,
  title={Fuseformer: Fusing fine-grained information in transformers for video inpainting},
  author={Liu, Rui and Deng, Hanming and Huang, Yangyi and Shi, Xiaoyu and Lu, Lewei and Sun, Wenxiu and Wang, Xiaogang and Dai, Jifeng and Li, Hongsheng},
  booktitle={Proceedings of the IEEE/CVF international conference on computer vision},
  pages={14040--14049},
  year={2021}
}

@article{liu2021decoupled,
  title={Decoupled spatial-temporal transformer for video inpainting},
  author={Liu, Rui and Deng, Hanming and Huang, Yangyi and Shi, Xiaoyu and Lu, Lewei and Sun, Wenxiu and Wang, Xiaogang and Dai, Jifeng and Li, Hongsheng},
  journal={arXiv preprint arXiv:2104.06637},
  year={2021}
}

@inproceedings{cai2022devit,
  title={Devit: Deformed vision transformers in video inpainting},
  author={Cai, Jiayin and Li, Changlin and Tao, Xin and Yuan, Chun and Tai, Yu-Wing},
  booktitle={Proceedings of the 30th ACM international conference on multimedia},
  pages={779--789},
  year={2022}
}

@inproceedings{zhou2023propainter,
  title={Propainter: Improving propagation and transformer for video inpainting},
  author={Zhou, Shangchen and Li, Chongyi and Chan, Kelvin CK and Loy, Chen Change},
  booktitle={Proceedings of the IEEE/CVF international conference on computer vision},
  pages={10477--10486},
  year={2023}
}

@article{guo2023animatediff,
  title={Animatediff: Animate your personalized text-to-image diffusion models without specific tuning},
  author={Guo, Yuwei and Yang, Ceyuan and Rao, Anyi and Liang, Zhengyang and Wang, Yaohui and Qiao, Yu and Agrawala, Maneesh and Lin, Dahua and Dai, Bo},
  journal={arXiv preprint arXiv:2307.04725},
  year={2023}
}

@inproceedings{zhang2024avid,
  title={Avid: Any-length video inpainting with diffusion model},
  author={Zhang, Zhixing and Wu, Bichen and Wang, Xiaoyan and Luo, Yaqiao and Zhang, Luxin and Zhao, Yinan and Vajda, Peter and Metaxas, Dimitris and Yu, Licheng},
  booktitle={Proceedings of the IEEE/CVF conference on computer vision and pattern recognition},
  pages={7162--7172},
  year={2024}
}

@article{green2024semantically,
  title={Semantically consistent video inpainting with conditional diffusion models},
  author={Green, Dylan and Harvey, William and Naderiparizi, Saeid and Niedoba, Matthew and Liu, Yunpeng and Liang, Xiaoxuan and Lavington, Jonathan and Zhang, Ke and Lioutas, Vasileios and Dabiri, Setareh and others},
  journal={arXiv preprint arXiv:2405.00251},
  year={2024}
}

@inproceedings{zi2025cococo,
  title={Cococo: Improving text-guided video inpainting for better consistency, controllability and compatibility},
  author={Zi, Bojia and Zhao, Shihao and Qi, Xianbiao and Wang, Jianan and Shi, Yukai and Chen, Qianyu and Liang, Bin and Xiao, Rong and Wong, Kam-Fai and Zhang, Lei},
  booktitle={Proceedings of the AAAI Conference on Artificial Intelligence},
  volume={39},
  number={10},
  pages={11067--11076},
  year={2025}
}

@inproceedings{bian2025videopainter,
  title={Videopainter: Any-length video inpainting and editing with plug-and-play context control},
  author={Bian, Yuxuan and Zhang, Zhaoyang and Ju, Xuan and Cao, Mingdeng and Xie, Liangbin and Shan, Ying and Xu, Qiang},
  booktitle={Proceedings of the Special Interest Group on Computer Graphics and Interactive Techniques Conference Conference Papers},
  pages={1--12},
  year={2025}
}

@article{li2025diffueraser,
  title={Diffueraser: A diffusion model for video inpainting},
  author={Li, Xiaowen and Xue, Haolan and Ren, Peiran and Bo, Liefeng},
  journal={arXiv preprint arXiv:2501.10018},
  year={2025}
}

@inproceedings{lee2025video,
  title={Video diffusion models are strong video inpainter},
  author={Lee, Minhyeok and Cho, Suhwan and Shin, Chajin and Lee, Jungho and Yang, Sunghun and Lee, Sangyoun},
  booktitle={Proceedings of the AAAI Conference on Artificial Intelligence},
  volume={39},
  number={4},
  pages={4526--4533},
  year={2025}
}

@inproceedings{wan2025unipaint,
  title={Unipaint: Unified space-time video inpainting via mixture-of-experts},
  author={Wan, Zhen and Qi, Chenyang and Liu, Zhiheng and Gui, Tao and Ma, Yue},
  booktitle={Proceedings of the IEEE/CVF International Conference on Computer Vision},
  pages={1861--1871},
  year={2025}
}

@inproceedings{guo2025keyframe,
  title={Keyframe-Guided Creative Video Inpainting},
  author={Guo, Yuwei and Yang, Ceyuan and Rao, Anyi and Meng, Chenlin and Bar-Tal, Omer and Ding, Shuangrui and Agrawala, Maneesh and Lin, Dahua and Dai, Bo},
  booktitle={Proceedings of the Computer Vision and Pattern Recognition Conference},
  pages={13009--13020},
  year={2025}
}

@inproceedings{hu2022make,
  title={Make it move: controllable image-to-video generation with text descriptions},
  author={Hu, Yaosi and Luo, Chong and Chen, Zhenzhong},
  booktitle={Proceedings of the IEEE/CVF Conference on Computer Vision and Pattern Recognition},
  pages={18219--18228},
  year={2022}
}

@article{namekata2024sg,
  title={Sg-i2v: Self-guided trajectory control in image-to-video generation},
  author={Namekata, Koichi and Bahmani, Sherwin and Wu, Ziyi and Kant, Yash and Gilitschenski, Igor and Lindell, David B},
  journal={arXiv preprint arXiv:2411.04989},
  year={2024}
}

@article{wang2024generative,
  title={Generative inbetweening: Adapting image-to-video models for keyframe interpolation},
  author={Wang, Xiaojuan and Zhou, Boyang and Curless, Brian and Kemelmacher-Shlizerman, Ira and Holynski, Aleksander and Seitz, Steven M},
  journal={arXiv preprint arXiv:2408.15239},
  year={2024}
}

@article{jin2024pyramidal,
  title={Pyramidal flow matching for efficient video generative modeling},
  author={Jin, Yang and Sun, Zhicheng and Li, Ningyuan and Xu, Kun and Jiang, Hao and Zhuang, Nan and Huang, Quzhe and Song, Yang and Mu, Yadong and Lin, Zhouchen},
  journal={arXiv preprint arXiv:2410.05954},
  year={2024}
}

@article{li2025anyi2v,
  title={AnyI2V: Animating Any Conditional Image with Motion Control},
  author={Li, Ziye and Luo, Hao and Shuai, Xincheng and Ding, Henghui},
  journal={arXiv preprint arXiv:2507.02857},
  year={2025}
}

@inproceedings{yang2025versatile,
  title={Versatile transition generation with image-to-video diffusion},
  author={Yang, Zuhao and Zhang, Jiahui and Yu, Yingchen and Lu, Shijian and Bai, Song},
  booktitle={Proceedings of the IEEE/CVF International Conference on Computer Vision},
  pages={16981--16990},
  year={2025}
}

@inproceedings{wang2025tip,
  title={Tip-i2v: A million-scale real text and image prompt dataset for image-to-video generation},
  author={Wang, Wenhao and Yang, Yi},
  booktitle={Proceedings of the IEEE/CVF International Conference on Computer Vision},
  pages={14898--14908},
  year={2025}
}

@inproceedings{shi2025motionstone,
  title={Motionstone: Decoupled motion intensity modulation with diffusion transformer for image-to-video generation},
  author={Shi, Shuwei and Gong, Biao and Chen, Xi and Zheng, Dandan and Tan, Shuai and Yang, Zizheng and Li, Yuyuan and He, Jingwen and Zheng, Kecheng and Chen, Jingdong and others},
  booktitle={Proceedings of the Computer Vision and Pattern Recognition Conference},
  pages={22864--22874},
  year={2025}
}

@inproceedings{zhang2025motionpro,
  title={MotionPro: A Precise Motion Controller for Image-to-Video Generation},
  author={Zhang, Zhongwei and Long, Fuchen and Qiu, Zhaofan and Pan, Yingwei and Liu, Wu and Yao, Ting and Mei, Tao},
  booktitle={Proceedings of the Computer Vision and Pattern Recognition Conference},
  pages={27957--27967},
  year={2025}
}

@inproceedings{yariv2025through,
  title={Through-The-Mask: Mask-based Motion Trajectories for Image-to-Video Generation},
  author={Yariv, Guy and Kirstain, Yuval and Zohar, Amit and Sheynin, Shelly and Taigman, Yaniv and Adi, Yossi and Benaim, Sagie and Polyak, Adam},
  booktitle={Proceedings of the Computer Vision and Pattern Recognition Conference},
  pages={18198--18208},
  year={2025}
}

@inproceedings{liu2025generative,
  title={Generative video propagation},
  author={Liu, Shaoteng and Wang, Tianyu and Wang, Jui-Hsien and Liu, Qing and Zhang, Zhifei and Lee, Joon-Young and Li, Yijun and Yu, Bei and Lin, Zhe and Kim, Soo Ye and others},
  booktitle={Proceedings of the Computer Vision and Pattern Recognition Conference},
  pages={17712--17722},
  year={2025}
}

@misc{tu2025videoanydoorhighfidelityvideoobject,
      title={VideoAnydoor: High-fidelity Video Object Insertion with Precise Motion Control}, 
      author={Yuanpeng Tu and Hao Luo and Xi Chen and Sihui Ji and Xiang Bai and Hengshuang Zhao},
      year={2025},
      journal={arXiv preprint arXiv:2501.01427}, 
}

@article{he2023animate,
  title={Animate-a-story: Storytelling with retrieval-augmented video generation},
  author={He, Yingqing and Xia, Menghan and Chen, Haoxin and Cun, Xiaodong and Gong, Yuan and Xing, Jinbo and Zhang, Yong and Wang, Xintao and Weng, Chao and Shan, Ying and others},
  journal={arXiv preprint arXiv:2307.06940},
  year={2023}
}

@article{jiang2025vace,
  title={Vace: All-in-one video creation and editing},
  author={Jiang, Zeyinzi and Han, Zhen and Mao, Chaojie and Zhang, Jingfeng and Pan, Yulin and Liu, Yu},
  journal={arXiv preprint arXiv:2503.07598},
  year={2025}
}

@inproceedings{wang2025videodirector,
  title={Videodirector: Precise video editing via text-to-video models},
  author={Wang, Yukun and Wang, Longguang and Ma, Zhiyuan and Hu, Qibin and Xu, Kai and Guo, Yulan},
  booktitle={Proceedings of the Computer Vision and Pattern Recognition Conference},
  pages={2589--2598},
  year={2025}
}

@article{wei2025univideo,
  title={Univideo: Unified understanding, generation, and editing for videos},
  author={Wei, Cong and Liu, Quande and Ye, Zixuan and Wang, Qiulin and Wang, Xintao and Wan, Pengfei and Gai, Kun and Chen, Wenhu},
  journal={arXiv preprint arXiv:2510.08377},
  year={2025}
}

@article{cheng2025wan,
  title={Wan-animate: Unified character animation and replacement with holistic replication},
  author={Cheng, Gang and Gao, Xin and Hu, Li and Hu, Siqi and Huang, Mingyang and Ji, Chaonan and Li, Ju and Meng, Dechao and Qi, Jinwei and Qiao, Penchong and others},
  journal={arXiv preprint arXiv:2509.14055},
  year={2025}
}

@article{chen2025contextflow,
  title={ContextFlow: Training-Free Video Object Editing via Adaptive Context Enrichment},
  author={Chen, Yiyang and He, Xuanhua and Ma, Xiujun and Ma, Yue},
  journal={arXiv preprint arXiv:2509.17818},
  year={2025}
}

@article{liu2025step1x,
  title={Step1x-edit: A practical framework for general image editing},
  author={Liu, Shiyu and Han, Yucheng and Xing, Peng and Yin, Fukun and Wang, Rui and Cheng, Wei and Liao, Jiaqi and Wang, Yingming and Fu, Honghao and Han, Chunrui and others},
  journal={arXiv preprint arXiv:2504.17761},
  year={2025}
}

@article{cai2025z,
  title={Z-Image: An Efficient Image Generation Foundation Model with Single-Stream Diffusion Transformer},
  author={Cai, Huanqia and Cao, Sihan and Du, Ruoyi and Gao, Peng and Hoi, Steven and Huang, Shijie and Hou, Zhaohui and Jiang, Dengyang and Jin, Xin and Li, Liangchen and others},
  journal={arXiv preprint arXiv:2511.22699},
  year={2025}
}

@article{wu2025qwen,
  title={Qwen-image technical report},
  author={Wu, Chenfei and Li, Jiahao and Zhou, Jingren and Lin, Junyang and Gao, Kaiyuan and Yan, Kun and Yin, Sheng-ming and Bai, Shuai and Xu, Xiao and Chen, Yilei and others},
  journal={arXiv preprint arXiv:2508.02324},
  year={2025}
}

@article{liu2024place,
  title={Place anything into any video},
  author={Liu, Ziling and Yang, Jinyu and Gao, Mingqi and Zheng, Feng},
  journal={arXiv preprint arXiv:2402.14316},
  year={2024}
}

@article{bai2024anything,
  title={Anything in any scene: Photorealistic video object insertion},
  author={Bai, Chen and Shao, Zeman and Zhang, Guoxiang and Liang, Di and Yang, Jie and Zhang, Zhuorui and Guo, Yujian and Zhong, Chengzhang and Qiu, Yiqiao and Wang, Zhendong and others},
  journal={arXiv preprint arXiv:2401.17509},
  year={2024}
}

@article{jin2025insertanywhere,
  title={InsertAnywhere: Bridging 4D Scene Geometry and Diffusion Models for Realistic Video Object Insertion},
  author={Jin, Hoiyeong and Jang, Hyojin and Kim, Jeongho and Hyung, Junha and Kim, Kinam and Kim, Dongjin and Choi, Huijin and Kim, Hyeonji and Choo, Jaegul},
  journal={arXiv preprint arXiv:2512.17504},
  year={2025}
}

@inproceedings{zhang2025scaling,
  title={Scaling in-the-wild training for diffusion-based illumination harmonization and editing by imposing consistent light transport},
  author={Zhang, Lvmin and Rao, Anyi and Agrawala, Maneesh},
  booktitle={The Thirteenth International Conference on Learning Representations},
  year={2025}
}

@article{miao2025rose,
  title={Rose: Remove objects with side effects in videos},
  author={Miao, Chenxuan and Feng, Yutong and Zeng, Jianshu and Gao, Zixiang and Liu, Hantang and Yan, Yunfeng and Qi, Donglian and Chen, Xi and Wang, Bin and Zhao, Hengshuang},
  journal={arXiv preprint arXiv:2508.18633},
  year={2025}
}

@article{lin2025depth,
  title={Depth anything 3: Recovering the visual space from any views},
  author={Lin, Haotong and Chen, Sili and Liew, Junhao and Chen, Donny Y and Li, Zhenyu and Shi, Guang and Feng, Jiashi and Kang, Bingyi},
  journal={arXiv preprint arXiv:2511.10647},
  year={2025}
}

@article{MOSEv2,
    title={{MOSEv2}: A More Challenging Dataset for Video Object Segmentation in Complex Scenes},
    author={Ding, Henghui and Ying, Kaining and Liu, Chang and He, Shuting and Jiang, Xudong and Jiang, Yu-Gang and Torr, Philip HS and Bai, Song},
    journal={arXiv preprint arXiv:2508.05630},
    year={2025}
}

@inproceedings{MOSE,
  title={{MOSE}: A New Dataset for Video Object Segmentation in Complex Scenes},
  author={Ding, Henghui and Liu, Chang and He, Shuting and Jiang, Xudong and Torr, Philip HS and Bai, Song},
  booktitle={ICCV},
  year={2023}
}

@article{Caelles_arXiv_2019,
  author = {Sergi Caelles and Jordi Pont-Tuset and Federico Perazzi and Alberto Montes and Kevis-Kokitsi Maninis and Luc {Van Gool}},
  title = {The 2019 DAVIS Challenge on VOS: Unsupervised Multi-Object Segmentation},
  journal = {arXiv},
  year = {2019}
}

@article{Pont-Tuset_arXiv_2017,
  author = {Jordi Pont-Tuset and Federico Perazzi and Sergi Caelles and Pablo Arbel\'aez and Alexander Sorkine-Hornung and Luc {Van Gool}},
  title = {The 2017 DAVIS Challenge on Video Object Segmentation},
  journal = {arXiv:1704.00675},
  year = {2017}
}

@inproceedings{athar2023burst,
  title={Burst: A benchmark for unifying object recognition, segmentation and tracking in video},
  author={Athar, Ali and Luiten, Jonathon and Voigtlaender, Paul and Khurana, Tarasha and Dave, Achal and Leibe, Bastian and Ramanan, Deva},
  booktitle={Proceedings of the IEEE/CVF winter conference on applications of computer vision},
  pages={1674--1683},
  year={2023}
}

@inproceedings{skorokhodov2022stylegan,
  title={Stylegan-v: A continuous video generator with the price, image quality and perks of stylegan2},
  author={Skorokhodov, Ivan and Tulyakov, Sergey and Elhoseiny, Mohamed},
  booktitle={Proceedings of the IEEE/CVF conference on computer vision and pattern recognition},
  pages={3626--3636},
  year={2022}
}

@inproceedings{zhang2018unreasonable,
  title={The unreasonable effectiveness of deep features as a perceptual metric},
  author={Zhang, Richard and Isola, Phillip and Efros, Alexei A and Shechtman, Eli and Wang, Oliver},
  booktitle={Proceedings of the IEEE conference on computer vision and pattern recognition},
  pages={586--595},
  year={2018}
}

@article{wang2004image,
  title={Image quality assessment: from error visibility to structural similarity},
  author={Wang, Zhou and Bovik, Alan C and Sheikh, Hamid R and Simoncelli, Eero P},
  journal={IEEE transactions on image processing},
  volume={13},
  number={4},
  pages={600--612},
  year={2004},
  publisher={IEEE}
}

@book{jahne2005digital,
  title={Digital image processing},
  author={J{\"a}hne, Bernd},
  year={2005},
  publisher={Springer}
}

@misc{GoogleCloud2025BananaPro,
  author = {{Google Cloud}},
  title = {Nano Banana Pro: State-of-the-Art Image Generation for Enterprise},
  year = {2025},
  month = nov,
  howpublished = {Google Cloud Blog},
  url = {https://cloud.google.com/blog/products/ai-machine-learning/nano-banana-pro-available-for-enterprise}
}

@inproceedings{huang2024vbench,
  title={Vbench: Comprehensive benchmark suite for video generative models},
  author={Huang, Ziqi and He, Yinan and Yu, Jiashuo and Zhang, Fan and Si, Chenyang and Jiang, Yuming and Zhang, Yuanhan and Wu, Tianxing and Jin, Qingyang and Chanpaisit, Nattapol and others},
  booktitle={Proceedings of the IEEE/CVF Conference on Computer Vision and Pattern Recognition},
  pages={21807--21818},
  year={2024}
}

@article{he2025openve,
  title={OpenVE-3M: A Large-Scale High-Quality Dataset for Instruction-Guided Video Editing},
  author={He, Haoyang and Wang, Jie and Zhang, Jiangning and Xue, Zhucun and Bu, Xingyuan and Yang, Qiangpeng and Wen, Shilei and Xie, Lei},
  journal={arXiv preprint arXiv:2512.07826},
  year={2025}
}

@article{chen2025ivebench,
  title={Ivebench: Modern benchmark suite for instruction-guided video editing assessment},
  author={Chen, Yinan and Zhang, Jiangning and Hu, Teng and Zeng, Yuxiang and Xue, Zhucun and He, Qingdong and Wang, Chengjie and Liu, Yong and Hu, Xiaobin and Yan, Shuicheng},
  journal={arXiv preprint arXiv:2510.11647},
  year={2025}
}

@article{polyak2024movie,
  title={Movie gen: A cast of media foundation models},
  author={Polyak, Adam and Zohar, Amit and Brown, Andrew and Tjandra, Andros and Sinha, Animesh and Lee, Ann and Vyas, Apoorv and Shi, Bowen and Ma, Chih-Yao and Chuang, Ching-Yao and others},
  journal={arXiv preprint arXiv:2410.13720},
  year={2024}
}

@misc{moore2024video,
  author       = {Moore, Justine},
  title        = {Why 2023 Was AI Video’s Breakout Year, and What to Expect in 2024},
  year         = {2024},
  month        = {jan},
  publisher    = {Andreessen Horowitz},
  url          = {https://a16z.com/why-2023-was-ai-videos-breakout-year-and-what-to-expect-in-2024/},
  note         = {Accessed: 2024-05-20}
}

@misc{manovich2023generative,
  author       = {Manovich, Lev},
  title        = {The Rise of Generative Media},
  year         = {2023},
  howpublished = {Manovich.net},
  url          = {http://manovich.net/content/04-projects/166-visual-generative-media/visual_generative_media.pdf},
  note         = {From the collection "Artificial Aesthetics"}
}

@article{ao2023image,
  title={Image amodal completion: A survey},
  author={Ao, Jiayang and Ke, Qiuhong and Ehinger, Krista A},
  journal={Computer Vision and Image Understanding},
  volume={229},
  pages={103661},
  year={2023},
  publisher={Elsevier}
}

@inproceedings{ao2025open,
  title={Open-world amodal appearance completion},
  author={Ao, Jiayang and Jiang, Yanbei and Ke, Qiuhong and Ehinger, Krista A},
  booktitle={Proceedings of the Computer Vision and Pattern Recognition Conference},
  pages={6490--6499},
  year={2025}
}

@article{wu2026consid,
  title={ConsID-Gen: View-Consistent and Identity-Preserving Image-to-Video Generation},
  author={Wu, Mingyang and Mishra, Ashirbad and Dey, Soumik and Xing, Shuo and Ravipati, Naveen and Wu, Hansi and Li, Binbin and Tu, Zhengzhong},
  journal={arXiv preprint arXiv:2602.10113},
  year={2026}
}

@article{yu2026sparkvsr,
  title={SparkVSR: Interactive Video Super-Resolution via Sparse Keyframe Propagation},
  author={Yu, Jiongze and Gao, Xiangbo and Verlani, Pooja and Gadde, Akshay and Wang, Yilin and Adsumilli, Balu and Tu, Zhengzhong},
  journal={arXiv preprint arXiv:2603.16864},
  year={2026}
}

@article{hacohen2026ltx,
  title={LTX-2: Efficient Joint Audio-Visual Foundation Model},
  author={HaCohen, Yoav and Brazowski, Benny and Chiprut, Nisan and Bitterman, Yaki and Kvochko, Andrew and Berkowitz, Avishai and Shalem, Daniel and Lifschitz, Daphna and Moshe, Dudu and Porat, Eitan and others},
  journal={arXiv preprint arXiv:2601.03233},
  year={2026}
}

@article{kong2024hunyuanvideo,
  title={Hunyuanvideo: A systematic framework for large video generative models},
  author={Kong, Weijie and Tian, Qi and Zhang, Zijian and Min, Rox and Dai, Zuozhuo and Zhou, Jin and Xiong, Jiangfeng and Li, Xin and Wu, Bo and Zhang, Jianwei and others},
  journal={arXiv preprint arXiv:2412.03603},
  year={2024}
}

@article{gao2026viskoorbis,
  title={{Visko Orbis 1.0}: A Live Model for Real-Time Interactive Long Video Generation},
  author={Gao, Xiangbo and Yang, Siyuan and He, Ping and Wu, Mingyang and Wu, Yuheng and Zuo, Yushen and Yu, Jiongze and Cui, Ryan and Hua, Hongyuan and Ma, Devin and Jin, Xiao and Ruan, Yubo and Yin, Qing and Yang, Jie and Tu, Zhengzhong},
  journal={arXiv preprint arXiv:2607.26694},
  year={2026},
  url={https://arxiv.org/abs/2607.26694}
}

@article{gao2026vefxbench,
  title={{VEFX-Bench}: A Holistic Benchmark for Generic Video Editing and Visual Effects},
  author={Gao, Xiangbo and Jiang, Sicong and Liu, Bangya and Chen, Xinghao and Yang, Minglai and Yang, Siyuan and Wu, Mingyang and Yu, Jiongze and Zheng, Qi and Wang, Haozhi and Zhang, Jiayi and Yang, Jie and Wang, Zihan and Yin, Qing and Tu, Zhengzhong},
  journal={arXiv preprint arXiv:2604.16272},
  year={2026},
  url={https://arxiv.org/abs/2604.16272}
}

@article{li2026physicsaware,
  title={Physics-Aware Video Instance Removal Benchmark},
  author={Li, Zirui and Chen, Xinghao and Jiang, Lingyu and Hou, Dengzhe and Lin, Fangzhou and Yamada, Kazunori and Gao, Xiangbo and Tu, Zhengzhong},
  journal={arXiv preprint arXiv:2604.05898},
  year={2026},
  url={https://arxiv.org/abs/2604.05898}
}

@article{wu2026deltaforcing,
  title={Delta Forcing: Trust Region Steering for Interactive Autoregressive Video Generation},
  author={Wu, Yuheng and Gao, Xiangbo and Chen, Tianhao and Chen, Xinghao and Yin, Qing and Tu, Zhengzhong and Lee, Dongman},
  journal={arXiv preprint arXiv:2605.14382},
  year={2026},
  url={https://arxiv.org/abs/2605.14382}
}

@article{gao2026pulsemotion,
  title={The Pulse of Motion: Measuring Physical Frame Rate from Visual Dynamics},
  author={Gao, Xiangbo and Wu, Mingyang and Yang, Siyuan and Yu, Jiongze and Taghavi, Pardis and Lin, Fangzhou and Tu, Zhengzhong},
  journal={arXiv preprint arXiv:2603.14375},
  year={2026},
  url={https://arxiv.org/abs/2603.14375}
}

@inproceedings{wu2026vgbeconsistent,
  title={{VGBE 2026} Challenge on Image-to-Video Consistent Generation: Methods and Results},
  author={Wu, Mingyang and Yang, Siyuan and Xing, Shuo and Mishra, Ashirbad and Dey, Soumik and Gao, Xiangbo and Zhao, Jinyu and Yu, Jiongze and Lin, Fangzhou and Tu, Zhengzhong and Lei, Yu and Feng, Zelin and Zhou, Xiaopeng and Guo, Yujie and Liu, Fang and Li, Lingling and Du, Hongyang and Ye, Junjie and Cong, Xiaoyan and Li, Runhao and Ni, Jingcheng and Agarwal, Aman and Zhou, Zeqi and Li, Zekun and Balestriero, Randall and Wang, Yue and Dasgupta, Debajyoti and Mishra, Parth and Garg, Gaurang and Goel, Pranshu and Kaur, Sukh Manpreet and Meng, Siwei and Luo, Pu and Xu, Cong and Li, Yumei and Jiao, Licheng and Liu, Xu and Ma, Wenping},
  booktitle={Proceedings of the IEEE/CVF Conference on Computer Vision and Pattern Recognition Workshops},
  pages={5201--5213},
  year={2026},
  url={https://openaccess.thecvf.com/content/CVPR2026W/VGBE/html/Wu_VGBE_2026_Challenge_on_Image-to-Video_Consistent_Generation_Methods_and_Results_CVPRW_2026_paper.html}
}

@article{wang2026generativeaiad,
  title={Generative {AI} for Autonomous Driving: Frontiers and Opportunities},
  author={Wang, Yuping and Xing, Shuo and Can, Cui and Li, Renjie and Hua, Hongyuan and Tian, Kexin and Mo, Zhaobin and Gao, Xiangbo and Wu, Keshu and Zhou, Sulong and You, Hengxu and Peng, Juntong and Zhang, Junge and Wang, Zehao and Song, Rui and Yan, Mingxuan and Zimmer, Walter and Zhou, Xingcheng and Li, Peiran and Lu, Zhaohan and Chen, Chia-Ju and Huang, Yue and Rossi, Ryan A. and Sun, Lichao and Yu, Hongkai and Fan, Zhiwen and Yang, Frank Hao and Kang, Yuhao and Greer, Ross and Liu, Chenxi and Lee, Eun Hak and Di, Xuan and Ye, Xinyue and Ren, Liu and Knoll, Alois and Li, Xiaopeng and Ji, Shuiwang and Tomizuka, Masayoshi and Pavone, Marco and Yang, Tianbao and Du, Jing and Yang, Ming-Hsuan and Wei, Hua and Wang, Ziran and Zhou, Yang and Li, Jiachen and Tu, Zhengzhong},
  journal={ACM Computing Surveys},
  year={2026},
  doi={10.1145/3838726},
  url={https://doi.org/10.1145/3838726}
}

@inproceedings{li2025fourkfourgen,
  title={{4K4DGen}: Panoramic {4D} Generation at {4K} Resolution},
  author={Li, Renjie and Pan, Panwang and Yang, Bangbang and Xu, Dejia and Zhou, Shijie and Zhang, Xuanyang and Li, Zeming and Kadambi, Achuta and Wang, Zhangyang and Tu, Zhengzhong and Fan, Zhiwen},
  booktitle={International Conference on Learning Representations},
  year={2025},
  url={https://openreview.net/forum?id=qxRoo7ULCo}
}

@inproceedings{he2024cover,
  title={{COVER}: A Comprehensive Video Quality Evaluator},
  author={He, Chenlong and Zheng, Qi and Zhu, Ruoxi and Zeng, Xiaoyang and Fan, Yibo and Tu, Zhengzhong},
  booktitle={Proceedings of the IEEE/CVF Conference on Computer Vision and Pattern Recognition Workshops},
  pages={5799--5809},
  year={2024},
  url={https://openaccess.thecvf.com/content/CVPR2024W/AI4Streaming/html/He_COVER_A_Comprehensive_Video_Quality_Evaluator_CVPRW_2024_paper.html}
}
}


\end{document}